\documentclass[11pt]{article}

\usepackage[preprint]{acl}

\usepackage{iftex}
\usepackage{placeins}

\ifPDFTeX
  \usepackage[T1]{fontenc}
  \usepackage[utf8]{inputenc}
  \usepackage{times}
  \usepackage{latexsym}
\else
  \usepackage{newtxtext}
  \usepackage{newtxmath}
  \usepackage{fontspec}
\fi

\ifPDFTeX
  \usepackage[T1]{fontenc}

  \usepackage[utf8]{inputenc}
\fi

\usepackage{microtype}

\usepackage{inconsolata}
\ifPDFTeX\else
\fi

\usepackage{graphicx}
\usepackage{float}

\usepackage{tabularx}
\usepackage{subcaption}
\usepackage{enumitem}

\definecolor{tokblue}{RGB}{0,90,200}
\definecolor{tokgreen}{RGB}{0,150,0}
\definecolor{tokorange}{RGB}{230,120,20}
\definecolor{tokred}{RGB}{200,30,30}

\newcommand{\bl}[1]{\textcolor{tokblue}{#1}}
\newcommand{\g}[1]{\textcolor{tokgreen}{#1}}
\newcommand{\ora}[1]{\textcolor{tokorange}{#1}}

\usepackage{booktabs}
\usepackage{colortbl}
\usepackage{multirow}
\usepackage{lipsum}
\usepackage[normalem]{ulem}

\ifPDFTeX
  \newcommand{\htok}[1]{#1}
  \newcommand{\ditok}[1]{#1}
  \newcommand{\tatok}[1]{#1}
\else
  \usepackage{polyglossia}
  \setmainlanguage{english}
  \setotherlanguage{hebrew}
  \setotherlanguage{hindi}
  \setotherlanguage{tamil}
  \newfontfamily\hebrewfont{FreeSerif}[
    Extension = .ttf,
    Script=Hebrew,Scale=1.1,
    BoldFont = FreeSerifBold,
    ItalicFont = FreeSerifItalic,
    BoldItalicFont = FreeSerifBoldItalic,
  ]
  \newfontfamily\hindifont{FreeSerif}[
    Extension = .ttf,
    Script=Devanagari,Scale=1.1,
    BoldFont = FreeSerifBold,
    ItalicFont = FreeSerifItalic,
    BoldItalicFont = FreeSerifBoldItalic,
  ]
  \newfontfamily\tamilfont{FreeSerif}[
    Extension = .ttf,
    Script=Tamil,Scale=1.1,
    BoldFont = FreeSerifBold,
    ItalicFont = FreeSerifItalic,
    BoldItalicFont = FreeSerifBoldItalic,
  ]
  \newcommand{\htok}[1]{\texthebrew{#1}}
  \newcommand{\ditok}[1]{\texthindi{#1}}
  \newcommand{\tatok}[1]{\texttamil{#1}}
\fi

\newcommand{\seg}[2]{%
  \ifdim #1 pt = 0.0 pt  \cellcolor{white}#2%
  \else\ifdim #1 pt = 0.29 pt \cellcolor{green!4}#2%
  \else\ifdim #1 pt = 0.33 pt \cellcolor{green!6}#2%
  \else\ifdim #1 pt = 0.4 pt \cellcolor{green!10}#2%
  \else\ifdim #1 pt = 0.5 pt  \cellcolor{green!15}#2%
  \else\ifdim #1 pt = 0.67 pt \cellcolor{green!23}#2%
  \else\ifdim #1 pt = 0.8 pt \cellcolor{green!30}#2%
  \else\ifdim #1 pt = 1.0 pt  \cellcolor{green!50}#2%
  \else \cellcolor{green!40}#2%
  \fi\fi\fi\fi\fi\fi\fi\fi
}

\usepackage{colortbl}
\usepackage{xcolor}
\usepackage{array}
\usepackage{booktabs}

\definecolor{headergreen}{RGB}{0,100,0}

\title{What Tokens are Learned when Tokenization is Optimized Jointly with Language Modeling?}

\author{Saketh Reddy Vemula \\
  IIIT Hyderabad, India \\
  \texttt{saketh.vemula@research.iiit.ac.in}
  \\\And
  Parameswari Krishnamurthy \\
  IIIT Hyderabad, India \\
  \texttt{param.krishna@iiit.ac.in} \\
}

\begin{document}
\maketitle
\begin{abstract}

Tokenization is a fundamental component of language modeling pipelines. Despite its importance, it is often fixed, even though it significantly impacts model performance across languages. In this work, we analyze what tokens are learned when tokenization is jointly optimized with language modeling. We compare \textit{tokenizer-free} approaches such as \g{SSLM}s and \bl{H-Net}s with fixed tokenizers across 18 typologically and script-diverse languages. Our results show that joint optimization fundamentally alters token structure. \g{SSLM}s recover morphologically aligned and contextually efficient tokens, whereas \bl{H-Net}s prioritize byte-level efficiency, producing longer tokens with very low overlap with standard subword vocabularies. We further show that tokenization behavior varies across language typologies. Agglutinative languages exhibit more dynamic segmentation patterns while learning. Through downstream evaluation, with pretrained-then-finetuned BERT models, we find that \g{SSLM}-based pretokenization consistently reduces language modeling perplexity and achieves competitive downstream performance despite distinct vocabularies. Overall, \textit{tokenizer-free} approaches optimize for contextual and computational efficiency rather than strict morphological structure, resulting in fundamentally different yet effective vocabularies for downstream NLP.

\end{abstract}

\section{Introduction}

Tokenization is a fundamental preprocessing step in language models (LMs). In transformer-based LMs \cite{vaswani2023attentionneed}, it is often performed by learning a tokenizer through subword algorithms such as Byte-pair Encoding (\ora{BPE}, \citet{gage_BPE_1994, sennrich-etal-2016-neural}), Unigram Language Model (\ora{ULM}, \citet{kudo-2018-subword}), or WordPiece (\ora{WPC}, \citet{wordPiece}). These algorithms are built on simple statistical priors, such as the assumption that frequently co-occurring characters should be part of the same token \cite{sennrich-etal-2016-neural}. Despite their widespread adoption, it is unclear whether such approaches provide best performance, especially when working with languages across diverse morphological complexities, typologies, and scripts. This concern is further amplified from the standard \textit{tokenization-LM-detokenization} pipeline, where a tokenizer is learned independently of language model training and remains static. We refer to these as \textit{fixed-tokenizer} approaches. Assessing the best tokenizer for such LMs is both difficult and expensive, since the tokenizer remains fixed prior to training and cannot be modified thereafter.
\begin{figure}[t]
  \centering
  \includegraphics[width=\columnwidth]{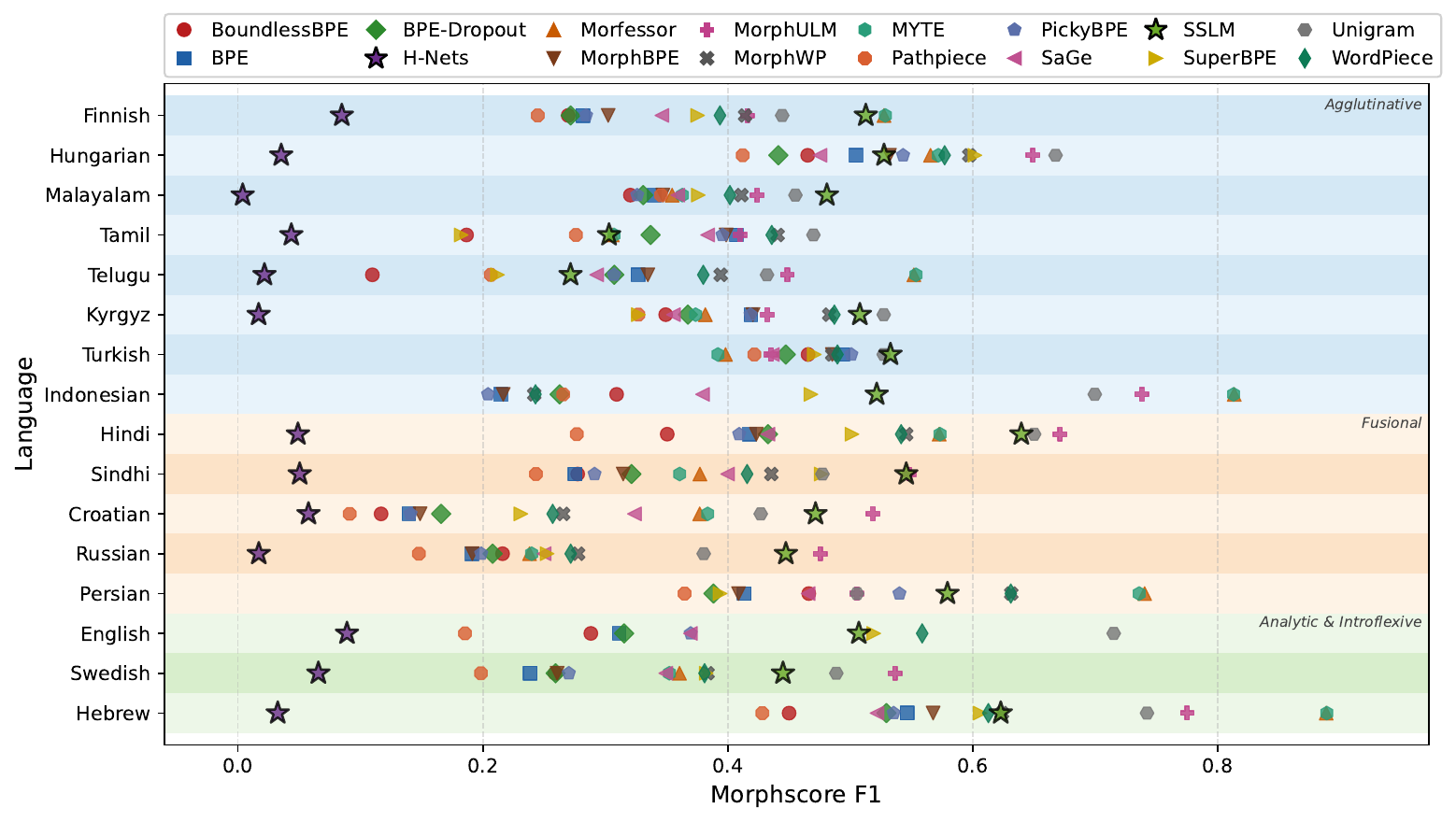}
  \caption{Morphological alignment evaluation of \textit{tokenizer-free} (\bl{H-Net}s, \g{SSLM}s) and fixed-tokenizer approaches. \g{SSLM}s show consistently higher alignment, whereas \bl{H-Net}s trade morphological alignment for longer, computationally efficient tokens.}
  \label{fig:typology_morphscore_f1}
\end{figure}

Various modifications and novel algorithms have been proposed to learn a tokenizer. This includes advances in pre-tokenization \cite{banerjee-bhattacharyya-2018-meaningless, hu2025entropydrivenpretokenizationbytepairencoding}, robustness \cite{kudo-richardson-2018-sentencepiece}, morphological alignment \cite{hofmann-etal-2022-embarrassingly, libovicky-helcl-2024-lexically, zhu-etal-2025-unsupervised-tree-tokenizer}, vocabulary pruning \cite{cognetta-etal-2024-vocabulary-pruning-analysis} or modifications \cite{chizhov-etal-2024-bpe}, and cross-token\footnote{Tokens that can span multiple words, for example ``by the way''.} approaches \cite{schmidt2025boundlessbytepairencoding, liu2025superbpespacetravellanguage}. However, no single approach can be concluded to be the ``best'', as conclusions depend on various factors such as: 1) languages chosen and their morphological properties, 2) experimental setup like model architecture, dataset size and other hyperparameters \cite{zhu-etal-2019-importance, poelman-etal-2025-confounding}.

Recently, new architectures have been proposed that overcome the aforementioned \textit{tokenization-LM-detokenization} pipeline. These are often referred to as \textit{tokenizer-free} LMs.
Architectures such as Dynamic Token Pooling Transformers \cite{nawrot-etal-2023-efficient-dynamic-token-pooling} and \bl{H-Net}s \cite{hwang2025dynamicchunkingendtoendhierarchical} incorporate data-dependent boundary prediction (i.e., tokenization) and byte-level language modeling into a single end-to-end pipeline. \bl{H-Net}s, in particular, provide a stable architecture. On the other hand, models such as subword segmental language model (\g{SSLM}, \citet{meyer-buys-2022-subword, meyer-buys-2025-learning}) use long short-term memory (LSTMs, \citet{long-short-term-memory-paper}) or transformers \cite{vaswani2023attentionneed} to marginalize and optimize over all possible segmentations. Despite being computationally heavy and unfeasible at large scale, \g{SSLM}s provide us with a joint architecture which we can utilize for analysis. These \textit{tokenizer-free} approaches avoid any statistical or heuristic priors, thus providing us with a tokenization that is jointly learned with the LM.


In this work, we investigate ``\textit{what tokens are learned when tokenization is optimized jointly with language modeling}.'' Our contributions:
\begin{enumerate}[leftmargin=*, labelindent=0pt, itemindent=0pt]
  \setlength\itemsep{0em}
  \item We analyze \textit{how} and \textit{what} tokens are learned through \textit{tokenizer-free} approaches, namely \g{SSLM} and \bl{H-Net}, by evaluating their intrinsic and linguistic properties across a diverse set of languages.
  \item We compare these approaches with \textit{fixed-tokenizer} methods, such as \ora{BPE} and \ora{ULM}, to assess \textit{how similar} their learned vocabularies are.
  \item We evaluate the impact of these tokenization approaches on downstream task performance to assess \textit{whether} jointly learned tokenizers offer improvements.
\end{enumerate} 

\section{Related Work}

Prior work analyzing tokens learned by \textit{tokenizer-free} LMs remains limited. Recently, \citet{meyer-buys-2025-learning} studied the learning dynamics of tokens learned by \g{SSLM}s \cite{meyer-buys-2022-subword}. They tracked subword learning dynamics from a linguistic perspective by evaluating morphological alignment, productivity, idiosyncrasy, and fertility, and identified distinct stages of subword learning. However, because their study focused on learning dynamics, the analysis was limited to three languages. Moreover, critical confounding factors such as dataset size and disparities across languages were not accounted for when drawing conclusions across languages.

On the other hand, many studies have analyzed fixed tokenizers and their modifications. These studies have largely focused on the impact of subword algorithms on downstream LM performance. For example, \citet{bostrom-durrett-2020-byte} and \citet{vemula-etal-2025-rethinking} compared different algorithms, showing that \ora{ULM} performs better on downstream tasks and recovers subwords that are more morphologically aligned. While, works such as \citet{uzan-etal-2024-greed} argued that inference strategy can matter as or more than the tokenizer construction algorithm.

\section{Methodology}

\subsection{Languages and Datasets}
\label{sec:languages_and_datasets}
\begin{table}[t]
\centering
\small
\begin{tabular}{p{0.96\columnwidth}}
\toprule
\textbf{Language Coverage} \\
\midrule
\textbf{Agglutinative:} Finnish, Hungarian, Malayalam, Tamil, Telugu, Kyrgyz, Turkish, Mongolian, Indonesian (\textbf{9 languages})\\
\addlinespace[2pt]
\textbf{Analytic \& Introflexive:} English, Swedish, Hebrew (\textbf{3 languages})\\
\addlinespace[2pt]
\textbf{Fusional:} Sanskrit, Hindi, Sindhi, Croatian, Russian, Persian (\textbf{6 languages}) \\
\bottomrule
\end{tabular}
\caption{List of typologically and script diverse languages included in this study. Detailed version of this table can be found in Table \ref{tab:language-coverage-table} of Appendix.}
\label{tab:language-coverage-short}
\end{table}
Languages vary widely in morphology, typology, and script, which may affect the best tokenization approach. We therefore study a broad, representative set of 18 languages across these diversity, listed in Table \ref{tab:language-coverage-short} (detailed version in Table \ref{tab:language-coverage-table} of Appendix).

An ideal comparison across languages requires parallel training and test data, which is difficult to obtain at this scale. We therefore use non-parallel monolingual corpora while carefully controlling the data source and size. To balance data sizes, we extract 250,000 English sentences and measure their size in bytes. Then, for other languages, we scale the data size according to \textit{byte-premiums}\footnote{Byte premium (BP) is the ratio of bytes required to encode a comparable amount of information or parallel text in one language relative to another language (typically designated as English).} (BP) \cite{arnett-etal-2024-a-bit-of-problem} and extract the desired number of sentences\footnote{This is necessary for our comparative analysis involving UTF-8 encoding, as it is known to produce disparities across languages with diverse scripts \cite{arnett-etal-2024-a-bit-of-problem}.}. We use WMT News Crawl corpora\footnote{\url{https://data.statmt.org/news-crawl/}} \cite{41880}, fixing the domain to News for most languages, falling back to the NLLB corpus \cite{nllbteam2022languageleftbehindscaling} where unavailable. Statistics of the resultant pretraining data along with available morphologically annotated data is listed in Table \ref{tab:language-dataset-stats} of the Appendix.

\subsection{Tokenization Approaches}
\label{sec:tokenization-approaches}
We include two distinct \textit{tokenizer-free} approaches from recent literatures:
\begin{itemize}[leftmargin=*, labelindent=0pt, itemindent=0pt]
  \setlength\itemsep{0em}
  \item Transformer-based version of subword segmental language model; \g{\textbf{SSLM}} \cite{meyer-buys-2025-learning}; that marginalize over all possible segmentations while jointly optimizing for language modeling.
  \item End-to-end \bl{\textbf{H-Net}}s \cite{hwang2025dynamicchunkingendtoendhierarchical} that jointly perform data-dependent boundary prediction along with byte-level language modeling.
\end{itemize}

To compare these with \textit{fixed-tokenizer} approaches, we include both standard and improved algorithms. Standard algorithms include \ora{\textbf{BPE}} \cite{sennrich-etal-2016-neural}, \ora{\textbf{ULM}} \cite{kudo-2018-subword}, and \ora{\textbf{WPC}} \cite{wordPiece}. We also study \ora{\textbf{SaGe}} \cite{yehezkel-pinter-2023-incorporating}, which prefers subword units occurring in fewer distinct contexts relative to their frequency, and unsupervised morphological tokenizers such as \ora{\textbf{Morfessor}} \cite{smit-etal-2014-morfessor}. We further include methods that modify inference (such as \ora{\textbf{BPE-dropout}} \cite{provilkov-etal-2020-bpe} and \ora{\textbf{PathPiece}} \cite{schmidt-etal-2024-tokenization-is-more-than-compression}), pre-tokenization (such as \ora{\textbf{MorphBPE}} \cite{banerjee-bhattacharyya-2018-meaningless}), encoding (such as \ora{\textbf{MYTE}} \cite{limisiewicz-etal-2024-myte}) and vocabulary construction (such as \ora{\textbf{PickyBPE}} \cite{chizhov-etal-2024-bpe}). We further include cross-token approaches such as \ora{\textbf{SuperBPE}} \cite{liu2025superbpespacetravellanguage} and \ora{\textbf{BoundlessBPE}} \cite{schmidt2025boundlessbytepairencoding}, which have been shown to be more efficient. Table \ref{tab:tokenizer_taxonomy} in Appendix describes these approaches and \S\ \ref{subsec:tokenizer_hyperparameters} lists their hyperparameters.

\subsection{Evaluation}
\label{sec:evaluation}

We evaluate both intrinsic properties and extrinsic performance of the mentioned tokenization approaches. We consider intrinsic metrics such as \textit{contextual exponence} and \textit{effective vocabulary size}, as well as linguistic metrics such as morphological alignment. Table \ref{tab:tokenizer-metrics} in the Appendix provides a comprehensive description of the metrics we evaluate. This broad intrinsic evaluation provides us with a meaningful assessment of the tokens produced by each tokenization approach.

To measure the impact of each tokenization approach on downstream performance, we pretrain BERT \cite{devlin-etal-2019-bert} models\footnote{Note that this requires treating \textit{tokenizer-free} approach as pretokenization. We discuss the reasons behind this in \S\ \ref{sec:downstream_evaluation}.} at a scale of 12M parameters\footnote{We perform an ablation study at different model scales and discuss the results in \S\ \ref{sec:downstream_evaluation} and in the Appendix.} and finetune them on tasks: Sentiment Analysis, POS Tagging, Named Entity Recognition (NER), and Dependency Parsing. We limit the downstream evaluation to three typologically distinct languages: English, Hindi, and Telugu; to keep the computational cost manageable.

\subsection{Experimental Setup}
We take a backward approach from the ideal experimental setup as noted by \citet{poelman-etal-2025-confounding},
allowing us to better isolate the effect of the tokenizer. We train \g{SSLM}s, \bl{H-Net}s, and \textit{fixed-tokenizer} algorithms on the train subset of the dataset. Training is monitored on the validation subset, and all tokenization approaches are evaluated on the test subset. 
Total parameters count of \bl{H-Net} models remains approximately 3M and that of \g{SSLM}s at 2M. This ensures that the ratio of token-to-parameter count is at least 2 across languages (see Table \ref{tab:language-dataset-stats}). While \bl{H-Net}s do not require any vocabulary for initialization, \g{SSLM}s require a fixed lexicon provided beforehand. We consider $10,000$ most frequent words for this, following \citet{meyer-buys-2025-learning}, and set maximum token length\footnote{Measured in terms of unicode character length, and not as length in raw UTF-8 bytes} as $5$. 
For downstream evaluation, we scrape a larger dataset containing 10M sentences from same data source; NewsCrawl \cite{41880}. This ensures a reasonable data size for 2M-30M parameter BERT models which we analyze. Detailed hyperparameters for each models are listed in the Appendix \ref{sec:hyperparameters}.

\begin{table*}[t]
\centering
\fontsize{7.5}{9}\selectfont   
\setlength{\tabcolsep}{4pt}
\setlength{\tabcolsep}{4pt}
\begin{tabular}{ccc ccc ccc ccc ccc ccc}
\toprule

\multicolumn{3}{c}{\textbf{English}} &
\multicolumn{3}{c}{\textbf{Hebrew}} &
\multicolumn{3}{c}{\textbf{Hindi}} &
\multicolumn{3}{c}{\textbf{Tamil}} &
\multicolumn{3}{c}{\textbf{Hungarian}} &
\multicolumn{3}{c}{\textbf{Indonesian}} \\
\midrule
\multicolumn{3}{c}{reaping} &
\multicolumn{3}{c}{\htok{משטרתיים}} &
\multicolumn{3}{c}{\ditok{कार्यकर्ताओं}} &
\multicolumn{3}{c}{\tatok{வகுத்து}} &
\multicolumn{3}{c}{brigádokkal} &
\multicolumn{3}{c}{ceritakan} \\

\multicolumn{3}{c}{reap ing} &
\multicolumn{3}{c}{\htok{משטרתי ים}} &
\multicolumn{3}{c}{\ditok{कार्यकर्ता ओं}} &
\multicolumn{3}{c}{\tatok{வகு த்து}} &
\multicolumn{3}{c}{brigád okkal} &
\multicolumn{3}{c}{cerita kan} \\

\midrule

\multicolumn{2}{c}{\textit{Segmentation}} & \textit{F1} &
\multicolumn{2}{c}{\textit{Segmentation}} & \textit{F1} &
\multicolumn{2}{c}{\textit{Segmentation}} & \textit{F1} &
\multicolumn{2}{c}{\textit{Segmentation}} & \textit{F1} &
\multicolumn{2}{c}{\textit{Segmentation}} & \textit{F1} &
\multicolumn{2}{c}{\textit{Segmentation}} & \textit{F1} \\

\midrule

\multicolumn{2}{c}{\seg{0.0}{rea ping}}          & 0   &
\multicolumn{2}{c}{\seg{0.0}{\htok{משט רתיים}}}  & 0   &
\multicolumn{2}{c}{\seg{0.0}{\ditok{का र्यकर ्ताओं}}}  & 0   &
\multicolumn{2}{c}{\seg{0.0}{\tatok{வக ுத்து}}}       & 0   &
\multicolumn{2}{c}{\seg{0.0}{bri g ádo kkal}}    & 0   &
\multicolumn{2}{c}{\seg{0.0}{ceri takan}}        & 0   \\

\multicolumn{2}{c}{\seg{0.67}{re ap ing}}         & 0.67 &
\multicolumn{2}{c}{\seg{0.5}{\htok{מ שטר תי ים}}} & 0.5  &
\multicolumn{2}{c}{\seg{0.67}{\ditok{कार्य कर्ता ओं}}}  & 0.67 &
\multicolumn{2}{c}{\seg{1.0}{\tatok{வகு த்து}}}         & 1    &
\multicolumn{2}{c}{\seg{0.5}{br igá d okkal}}     & 0.5  &
\multicolumn{2}{c}{\seg{0.67}{c erita kan}}       & 0.67 \\

\multicolumn{2}{c}{\seg{1.0}{reap ing}}           & 1    &
\multicolumn{2}{c}{\seg{0.67}{\htok{משטר תי ים}}} & 0.67 &
\multicolumn{2}{c}{\seg{0.67}{\ditok{कार्य कर्ता ओं}}}  & 0.67 &
\multicolumn{2}{c}{\seg{1.0}{\tatok{வகு த்து}}}         & 1    &
\multicolumn{2}{c}{\seg{0.0}{br igá dok kal}}     & 0    &
\multicolumn{2}{c}{\seg{0.67}{c erita kan}}       & 0.67 \\

\multicolumn{2}{c}{\seg{1.0}{reap ing}}           & 1    &
\multicolumn{2}{c}{\seg{0.5}{\htok{משטר ת י ים}}}  & 0.5  &
\multicolumn{2}{c}{\seg{0.67}{\ditok{कार्य कर्ता ओं}}}  & 0.67 &
\multicolumn{2}{c}{\seg{0.0}{\tatok{வ குத்த ு}}}        & 0    &
\multicolumn{2}{c}{\seg{0.67}{b rigád okkal}}     & 0.67 &
\multicolumn{2}{c}{\seg{0.67}{c erita kan}}       & 0.67 \\

\multicolumn{2}{c}{\seg{1.0}{reap ing}}           & 1    &
\multicolumn{2}{c}{\seg{0.5}{\htok{משטר ת י ים}}}  & 0.5  &
\multicolumn{2}{c}{\seg{0.67}{\ditok{कार्य कर्ता ओं}}}  & 0.67 &
\multicolumn{2}{c}{\seg{0.0}{\tatok{வ குத்த ு}}}        & 0    &
\multicolumn{2}{c}{\seg{0.67}{b rigád okkal}}     & 0.67 &
\multicolumn{2}{c}{\seg{0.5}{c erita k an}}        & 0.5  \\

\bottomrule
\end{tabular}
\caption{Evolution of segmentation of few candidate wordforms produced by \textit{tokenizer-free} approach \g{SSLM} with morphological alignment F1-scores. English, Hebrew, and Hindi are fusional or analytical languages, while Tamil, Hungarian, and Indonesian are agglutinative languages. More examples in Table \ref{tab:seg_f1_more} in Appendix.}
\label{tab:seg_f1}
\end{table*}

\section{Experiments \& Results}


\subsection{Q1: How are the tokens learned?}
\label{sec:learning_dynamics}

We analyze the learning dynamics of \g{SSLM}s similar to \citet{meyer-buys-2025-learning}, but with broader language coverage, script variation, and, more importantly, dataset sizes adjusted according to \textit{byte-premiums} \cite{arnett-etal-2024-a-bit-of-problem}. We evaluate and track morphological alignment against MorphScore \cite{arnett-bergen-2025-language, arnett2025evaluatingmorphologicalalignmenttokenizers} data, along with intrinsic properties such as effective vocabulary size and number of distinct neighbouring tokens, i.e., contextual exponence \cite{yehezkel-pinter-2023-incorporating}\footnote{Refer to Table \ref{tab:tokenizer-metrics} in Appendix for brief description of these metric.}.
Our results confirm the findings of \citet{meyer-buys-2025-learning} with inclusion of more diverse languages.

\subsubsection*{Morphological Alignment}
\begin{figure}[t]
  \centering
  \includegraphics[width=\columnwidth]{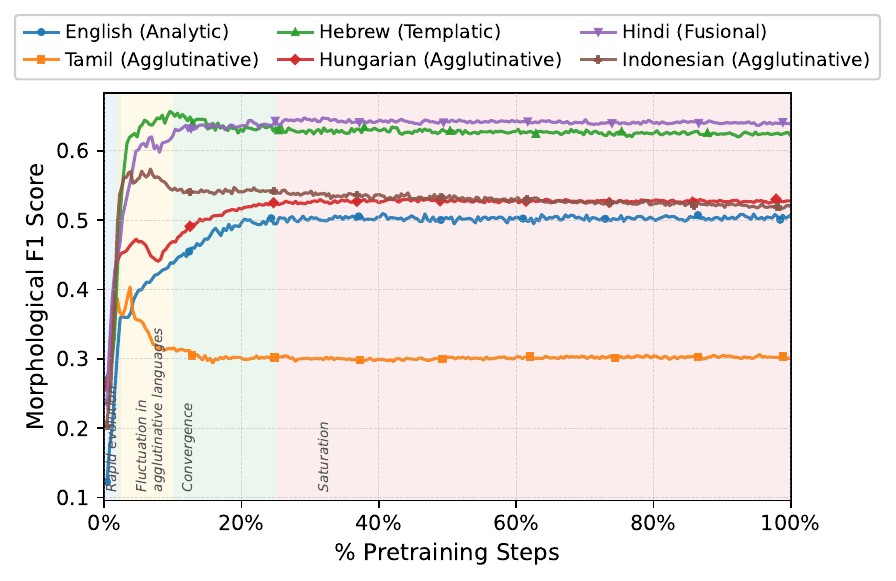}
  \caption{Evolution of morphological alignment (F1-score) of \textit{tokenizer-free} approach \g{SSLM} shows that agglutinative languages show larger fluctuations compared to other languages.}
  \label{fig:morphscore_dynamics}
\end{figure}

Figure \ref{fig:morphscore_dynamics} plots the variation in morphological alignment of tokens produced by \g{SSLM}, as its training progress, for a subset of languages; a similar plot containing all 18 languages is plotted in figure \ref{fig:morphscore_dynamics_all_languages} in Appendix. 

\textbf{Findings:} Across all languages, we observe four distinct phases in the morphscore dynamics: 1) \textbf{Rapid initial evolution} of morphemes, 2) \textbf{Fluctuation} for a brief period which is more consistent and pronounced in case of Agglutinative languages, 3) \textbf{Convergence} to a fixed alignment, and 4) \textbf{Saturation}, where alignment remains static implying convergence to a solution. Overall, higher alignment in Templatic and Fusional languages as compared to Agglutinative languages. 
This implies that a jointly learned tokenizer effectively identifies the root-and-pattern structures (as in Hebrew) and the inflectional suffixes (as in Hindi) very early in training. Despite being highly synthetic, Tamil converges to comparatively much lower F1-score ($\approx0.3$). Interestingly, the tokens reach mid-range alignment, but then regresses to a lower alignment. This is observed in both Tamil and Indonesian. Hungarian also shows similar trends like Tamil, but converges to higher alignment following another fluctuation. English, on the other hand, reaches a stable, mid-range alignment ($\approx0.5$) without any major fluctuations. This may reflect its relatively simple morphology. 

We identify and list candidate data-points from morphscore dataset (see Table \ref{tab:seg_f1}), which depicts the overall trend. For Analytic and Fusional languages, we observe how quickly the jointly learned tokenizer identifies the morphemes. In English, it instantly progresses from nonsensical split (``rea ping'', at step 2200) to a perfect morphologically aligned segment (``reap ing'', at step 2950). Similarly in Hindi, it converges to a stable segmentation (``\ditok{कार्य+कर्ता+ओं}'', at step 800) capturing the core semantic units. This implies that in languages where word structures are comparatively less complex, it converges on a stable ``lexicon'' of subwords. However, in Agglutinative languages such as Tamil, it reaches a perfect alignment (``\tatok{வகு+த்து}'', at step 450), but then converges on an inaccurate split (``\tatok{வ+குத்த+ ு}'', at step 7150), implying that the jointly learned tokenizer found it beneficial to trade-off morphological alignment in such languages.

\subsubsection*{Intrinsic Properties}
We perform similar analysis for intrinsic properties such as contextual exponence
and effective vocabulary size on test split.

\begin{figure}[t]
  \centering
  
  \begin{subfigure}{\columnwidth}
    \centering
    \includegraphics[width=\linewidth]{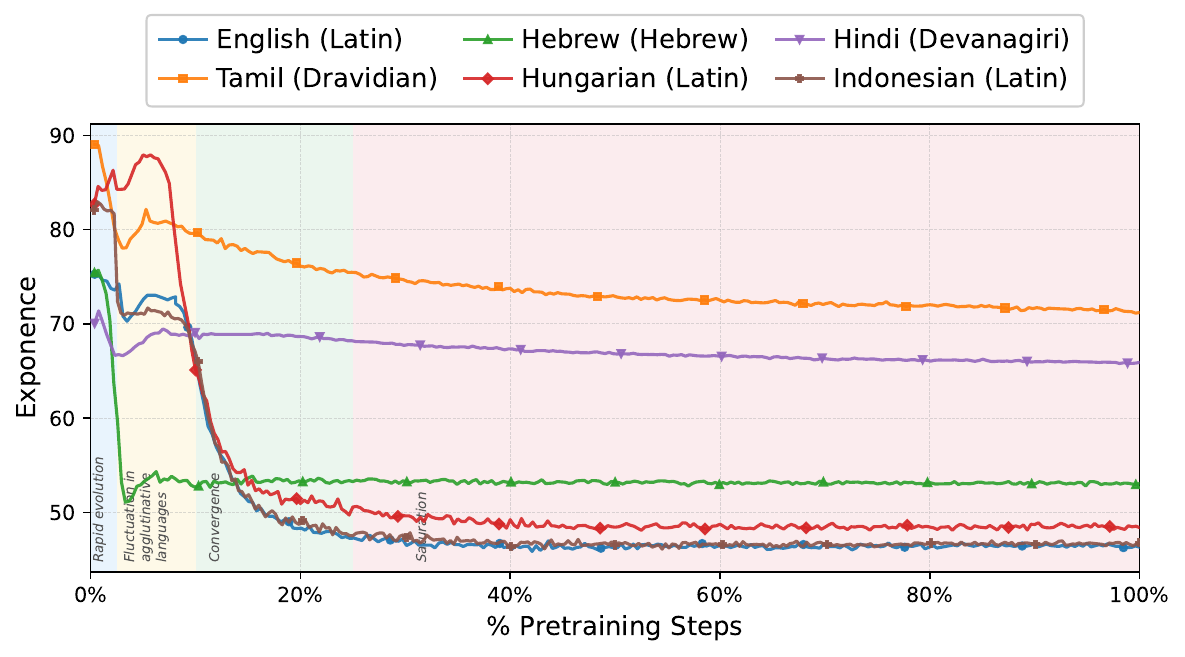}
    \caption{Contextual Exponence}
    \label{fig:exponence_dynamics}
  \end{subfigure}
  \begin{subfigure}{\columnwidth}
    \centering
    \includegraphics[width=\linewidth]{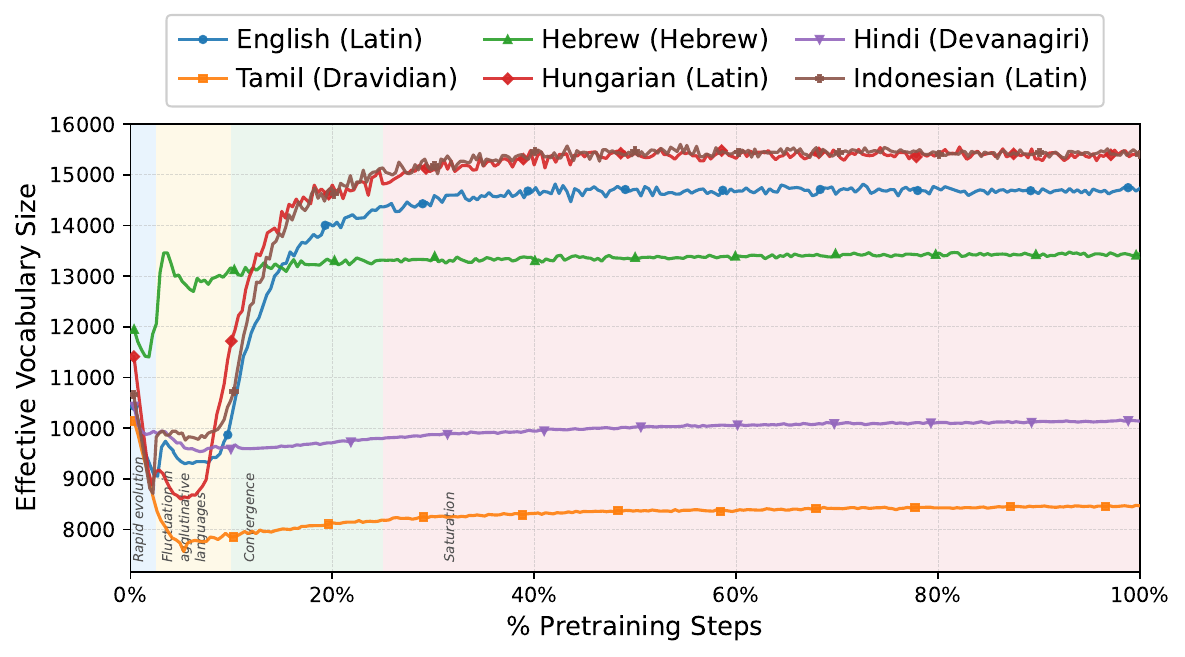}
    \caption{Effective Vocabulary size}
    \label{fig:vocab_size_dynamics}
  \end{subfigure}
  \caption{Evolution of number of distinct neighbouring tokens (i.e., contextual exponence) and effective vocabulary size of \textit{tokenizer-free} approach \g{SSLM} shows a script-wise similarity. Latin languages saturates at lower contextual exponence through higher effective vocabulary size.}
  \label{fig:combined_dynamics}
\end{figure}

\textbf{Findings:} Figure \ref{fig:exponence_dynamics} and \ref{fig:vocab_size_dynamics} reveal script-wise similarities in dynamics. Latin-script languages such as English, Hungarian, and Indonesian show largest effective vocabulary sizes, reaching between $14,000$ to $15,500$ unique tokens. This suggests that jointly learned tokenizer combines frequent character sequences into larger dedicated units to optimize contextual efficiency. We thereby see a steep drop in contextual exponence in these languages. 

Among the subset of languages here, non-Latin languages such as Tamil and Hindi show smaller effective vocabulary size and higher contextual exponence. This suggests that joint tokenizer prefers to learn a small, flexible set of fundamental morphemes rather than memorizing long, specific wordforms in such scripts. Dravidian-script languages such as Tamil combine many parts together, so one learned subword can still work with many different neighboring units because words often have many added affixes. This is observed through higher contextual exponence and lower vocabulary size. In contrast, Hebrew reaches its stable effective vocabulary size much earlier than other scripts. This shows again the ability of jointly learned tokenizers to quickly identify the root-and-pattern (introflexive) structures inherent in the script.

\subsection{Q2: What tokens are learned?}
\label{sec:token_properties}
In this section, we evaluate both \textit{tokenizer-free} approaches\footnote{We consider the tokens corresponding to the best checkpoint, i.e., the checkpoint corresponding to the least validation loss.}; \bl{H-Net}s and \g{SSLM}s; and compare them with \textit{fixed-tokenizer} approaches mentioned in \S\ \ref{sec:tokenization-approaches}. We focus on metrics listed in Table \ref{tab:tokenizer-metrics} of Appendix. \textbf{Note:} \bl{H-Net}s and \g{SSLM}s can, by design, produce fundamentally different types of tokens; \bl{H-Net}s can produce cross-tokens (or superwords) frequently, while \g{SSLM}s are bounded to produce only subwords. This is reflected in the further results and discussions.

\subsubsection*{Morphological Alignment}

\textbf{Findings:} As observed in figure \ref{fig:typology_morphscore_f1}, we identify a significant difference among tokens produced by the two \textit{tokenizer-free} approaches. \g{SSLM}s trade-off computational efficiency for high linguistic fidelity, while \bl{H-Net}s prioritize byte-level computational efficiency at the direct expense of morphological alignment. This can be observed by \bl{H-Net}'s low morphscore-F1 ranging below $0.1$ across all languages. On the other hand, joint optimization approach (i.e., \g{SSLM}) showcase significantly higher alignment ranging from $\approx0.26$ in Telugu to $\approx0.65$ in Hindi. Notably, \g{SSLM}s show consistently higher alignment as compared to popular subword algorithms such as \ora{BPE} or \ora{WPC}, even converging to the most aligned tokens in languages such as Malayalam and Turkish. However in other languages it tends to align less compared to approaches such as \ora{ULM}, \ora{MorphULM}, \ora{MYTE}, or \ora{Morfessor}. In Hebrew, despite such \textit{fixed-tokenizer} approaches providing nearly perfect alignment with morphology, \g{SSLM}s show comparatively lower alignment ($\approx0.63$). These findings suggest that the desirable morphological alignment of jointly learned tokenizer is indeed dependent on a language's morphological properties. 


\subsubsection*{Intrinsic Properties}

\begin{figure}[t]
  \centering
  
  \begin{subfigure}{\columnwidth}
    \centering
    \includegraphics[width=\columnwidth]{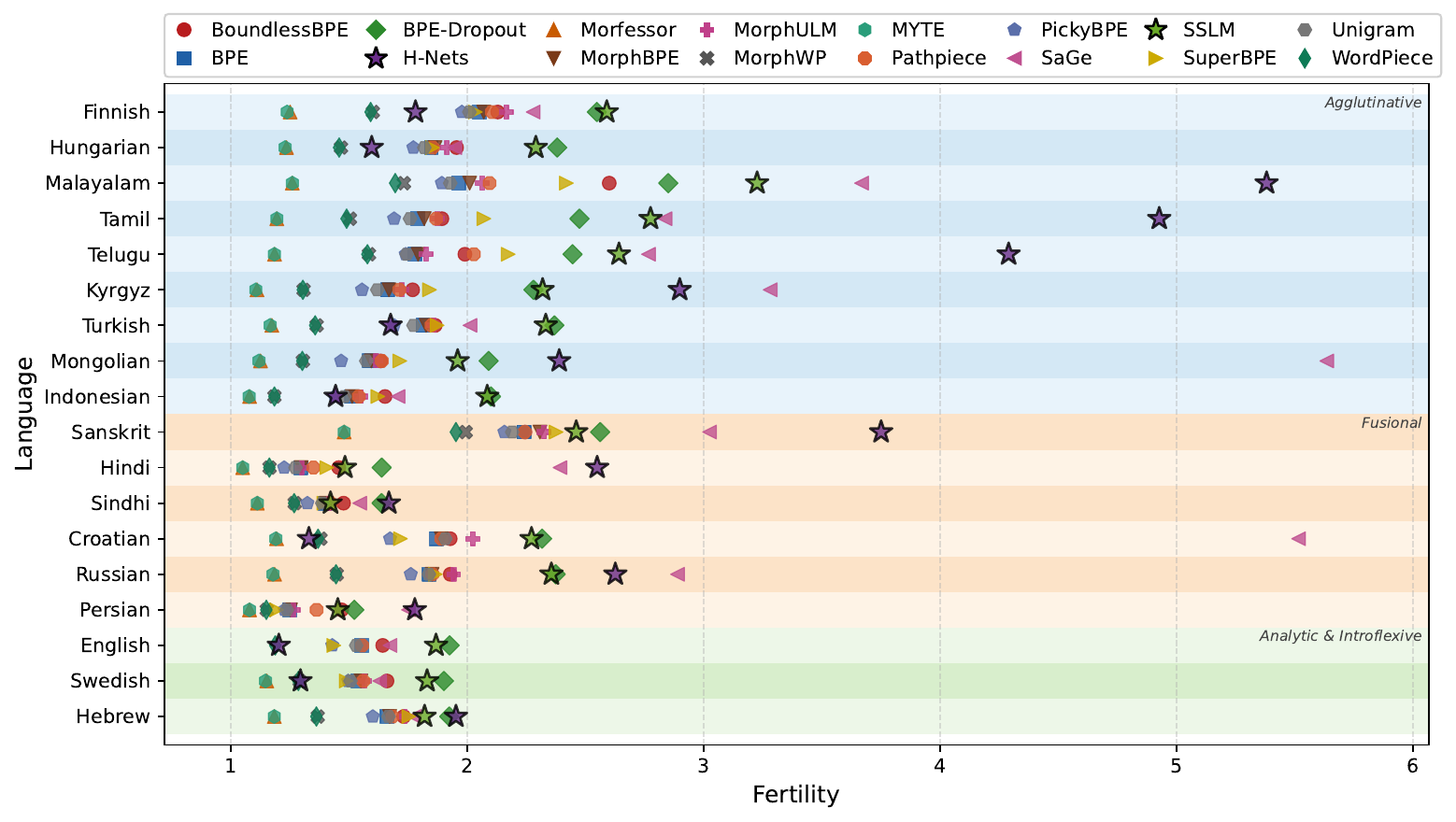}
    \caption{Fertility}
    \label{fig:typology_fertility}
  \end{subfigure}
  
  
  \begin{subfigure}{\columnwidth}
    \centering
    \includegraphics[width=\columnwidth]{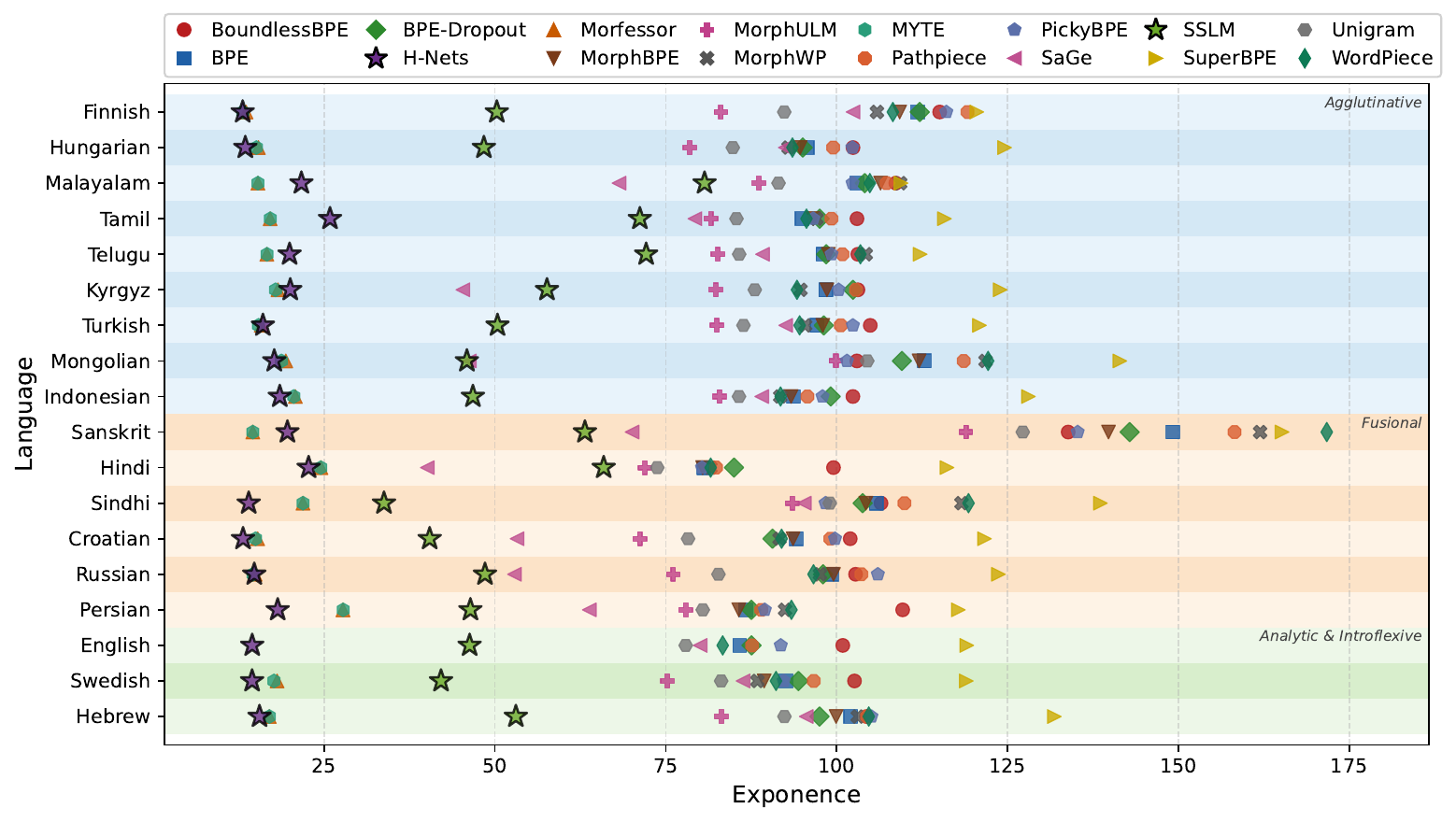}
    \caption{Contextual exponence}
    \label{fig:typology_exponence}
  \end{subfigure}

  \begin{subfigure}{\columnwidth}
    \centering
    \includegraphics[width=\columnwidth]{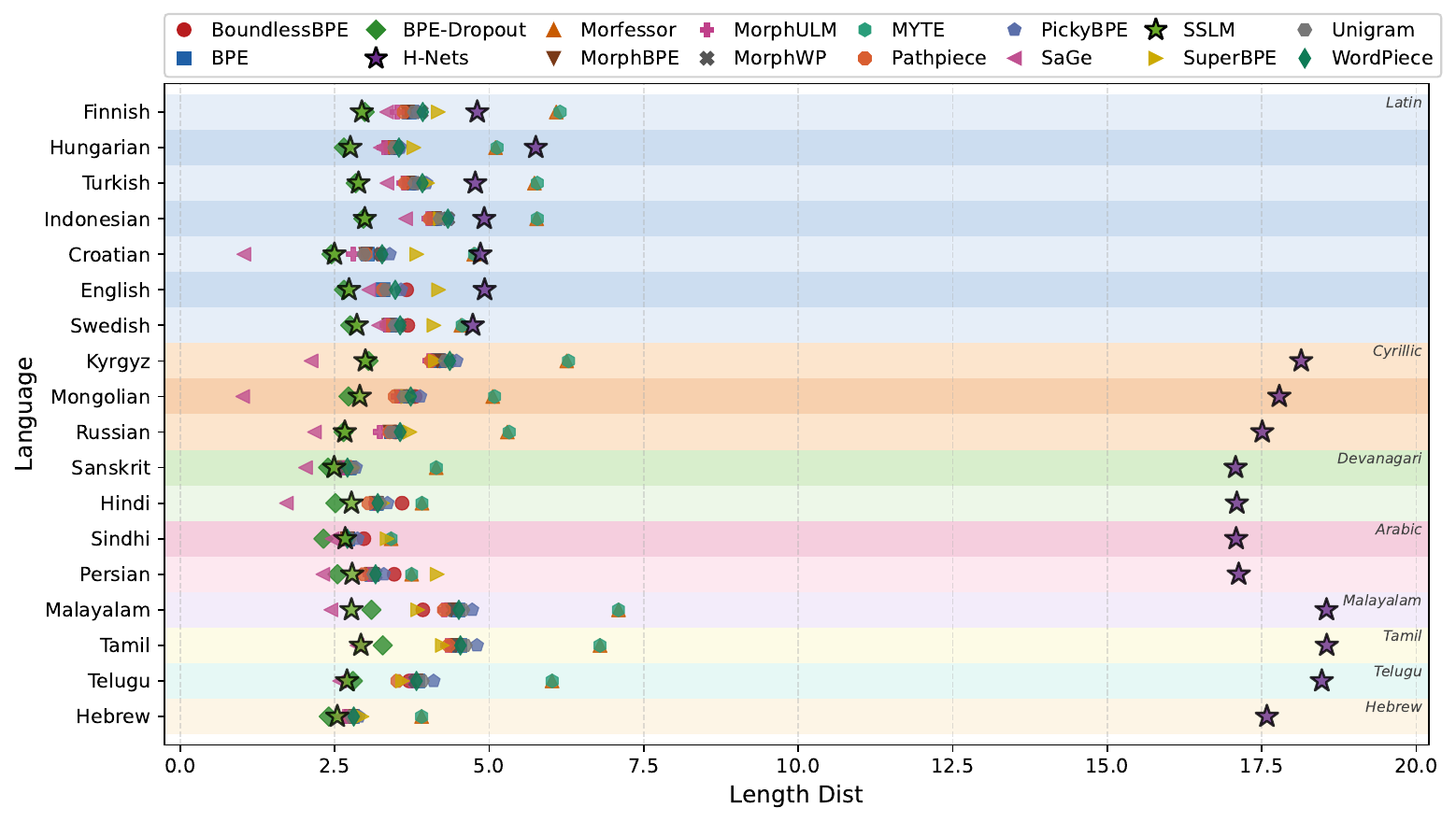}
    \caption{Mean Token Length}
    \label{fig:script_mean_token_length}
  \end{subfigure}

  \caption{Intrinsic evaluation (fertility, contextual exponence, and mean token length) of \textit{tokenizer-free} approaches (\g{SSLM}s and \bl{H-Net}s) and \textit{fixed-tokenizer} approaches. \textit{Tokenizer-free} approaches show comparatively higher fertility and lower contextual exponence. Mean token length of \bl{H-Net}s are significantly higher for non-Latin languages.}
  \label{fig:intrinsic_evaluation}
\end{figure}

\textbf{Findings:} As observed in figure \ref{fig:typology_fertility}, both \textit{tokenizer-free} approaches exhibit higher fertility than \textit{fixed-tokenizer} baselines, especially in agglutinative languages such as Malayalam, Tamil, and Telugu; this effect is strongest in \bl{H-Net}s ($\approx5.0$ to $\approx5.5$), while analytic or fusional languages remain below $\approx2.0$.

Across languages, \textit{tokenizer-free} approaches consistently yield lower contextual exponence than \textit{fixed-tokenizer} methods (see figure \ref{fig:typology_exponence}), with \g{SSLM} and \bl{H-Net} averaging less than $75$ neighbouring tokens. This likely arises because minimizing language-modeling loss favors units with more predictable grammatical roles and fewer valid neighbours, improving semantic quality and downstream performance \cite{yehezkel-pinter-2023-incorporating}.

\begin{figure*}[t]
    \centering

    \begin{subfigure}[b]{0.48\textwidth}
        \centering
        \includegraphics[width=\linewidth]{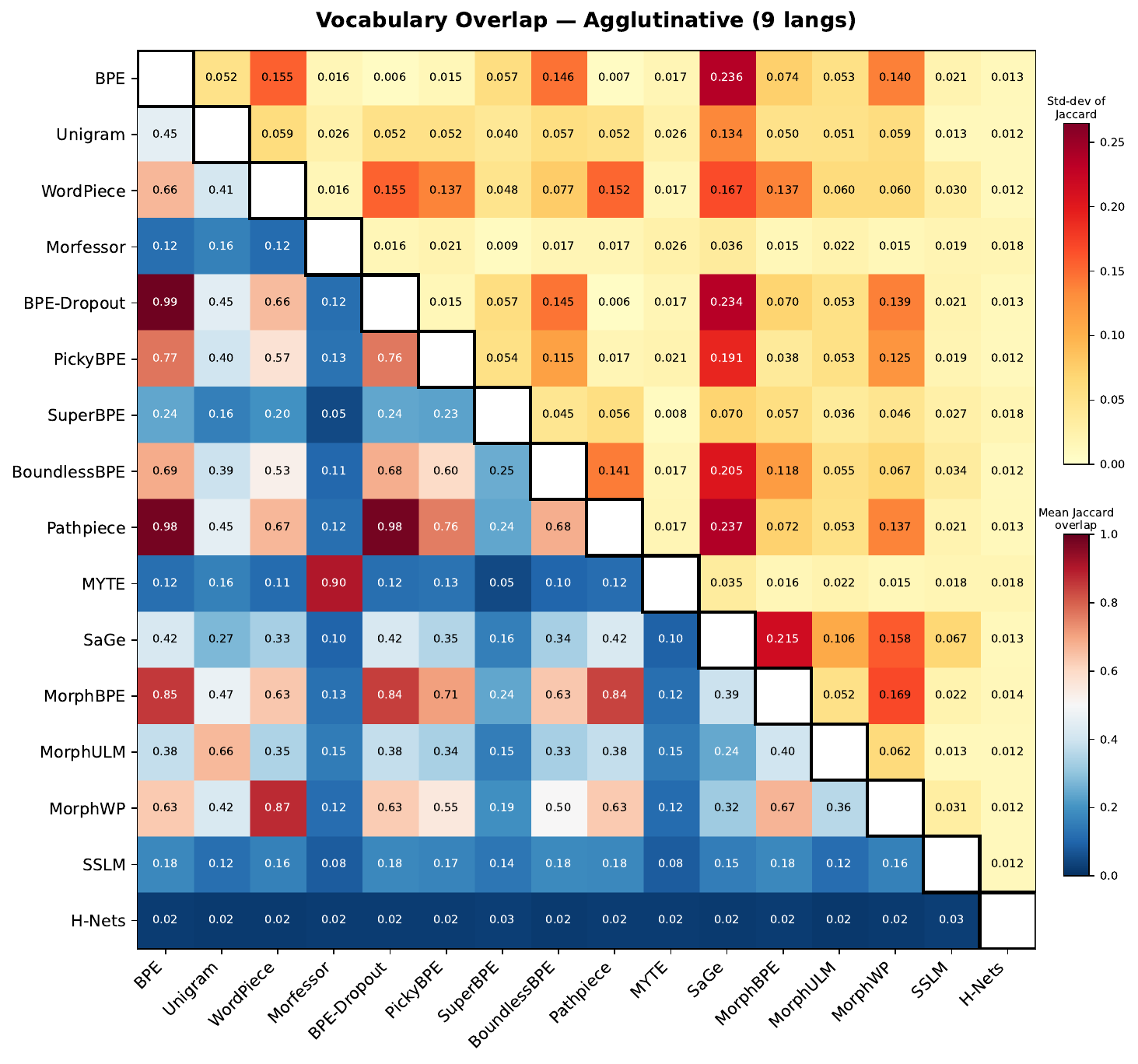}
        \caption{Agglutinative Languages}
        \label{fig:vocab_overlap_agglutinative}
    \end{subfigure}
    \hspace{0.01\textwidth}
    \begin{subfigure}[b]{0.48\textwidth}
        \centering
        \includegraphics[width=\linewidth]{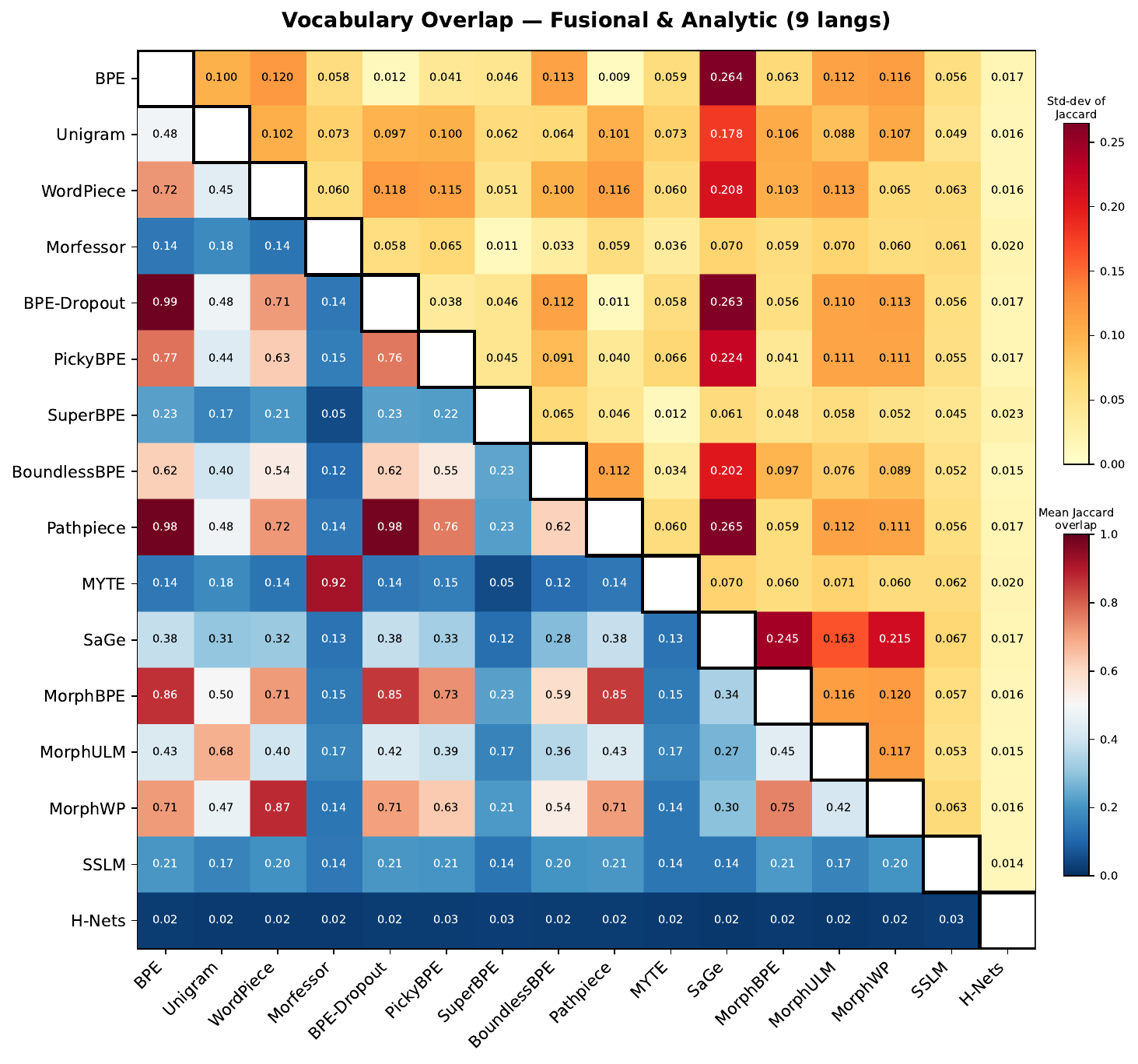}
        \caption{Analytic and Fusional Languages}
        \label{fig:vocab_overlap_analytic_fusional}
    \end{subfigure}

    \caption{Jaccard token overlap between \textit{tokenizer-free} approaches such as \g{SSLM} and \bl{H-Net}, and fixed-tokenizer approaches on the test split. The lower triangle of the heatmap denotes the Jaccard overlap, while the upper triangle denotes the standard deviation of the overlap.}
    \label{fig:vocab_overlap}
\end{figure*}

Interestingly, mean token length reveals strong divergence between \textit{tokenizer-free} approaches (see figure \ref{fig:script_mean_token_length}), especially across scripts. In non-Latin languages, \bl{H-Net}s produce exceptionally long tokens ($\approx17$ to $18.5$ characters), likely because either data-dependent boundary prediction favors long phrases to minimize byte-level loss or struggles to identify boundaries in complex scripts. In contrast, \g{SSLM}s and most fixed tokenizers maintain shorter, stable lengths ($\approx2.5$ to $5.0$ characters) which follows from its design of explicit maximum token-length constraints.

\subsection{Q3: How similar are the learned tokens?}
\label{sec:pairwise_token_overlap}

\begin{figure*}[t]
    \centering

    \begin{subfigure}[b]{0.30\textwidth}
        \centering
        \includegraphics[width=\linewidth]{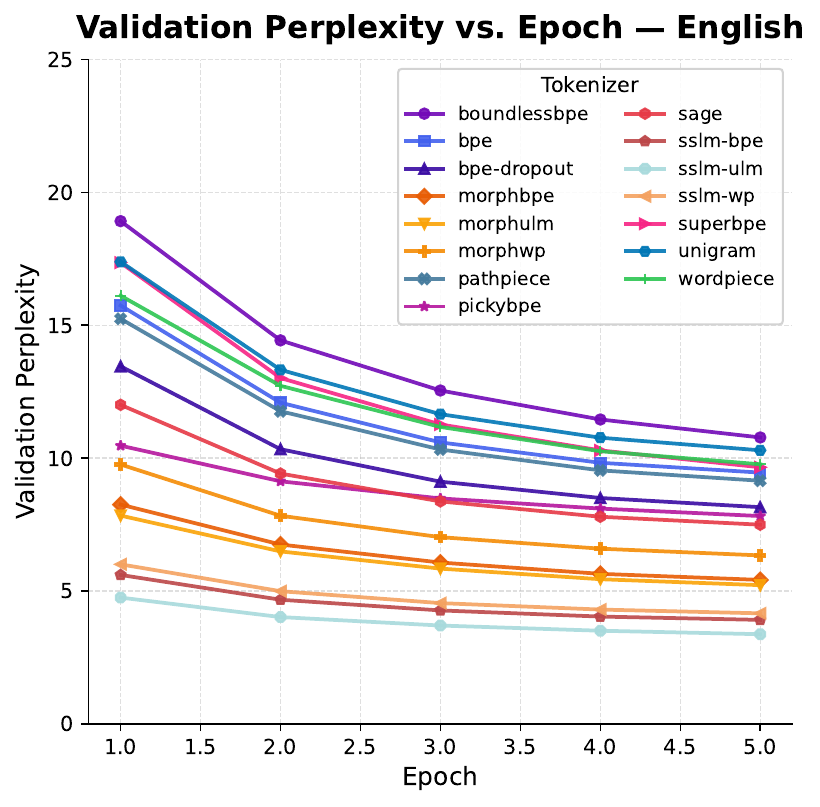}
        \caption{English (Analytic)}
        \label{fig:eval_perplexity_eng}
    \end{subfigure}
    \hspace{0.01\textwidth}
    \begin{subfigure}[b]{0.30\textwidth}
        \centering
        \includegraphics[width=\linewidth]{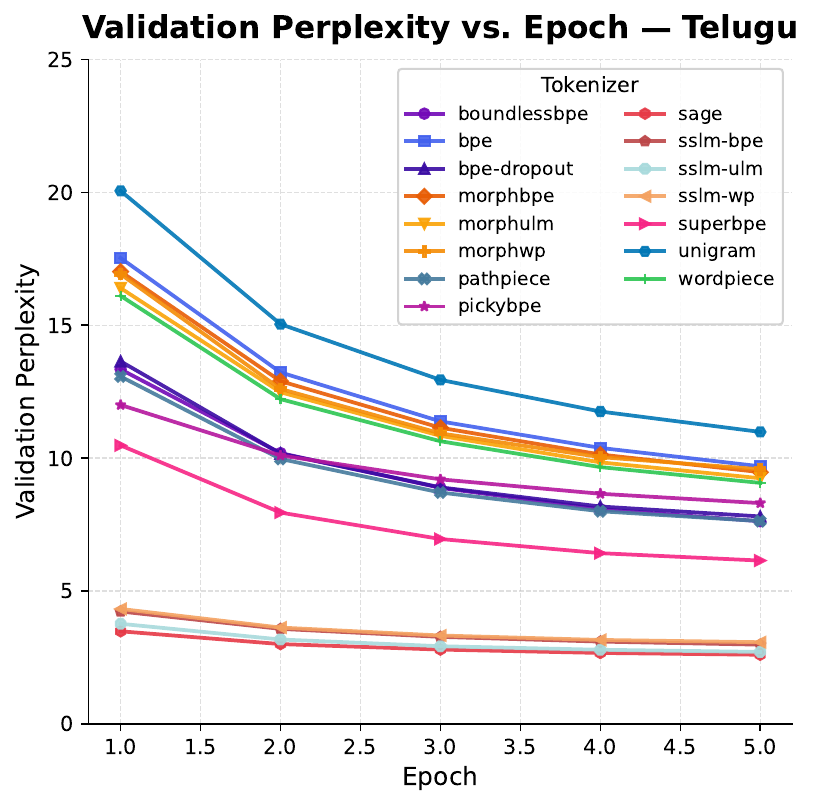}
        \caption{Telugu (Agglutinative)}
        \label{fig:eval_perplexity_tel}
    \end{subfigure}
    \hspace{0.01\textwidth}
    \begin{subfigure}[b]{0.30\textwidth}
        \centering
        \includegraphics[width=\linewidth]{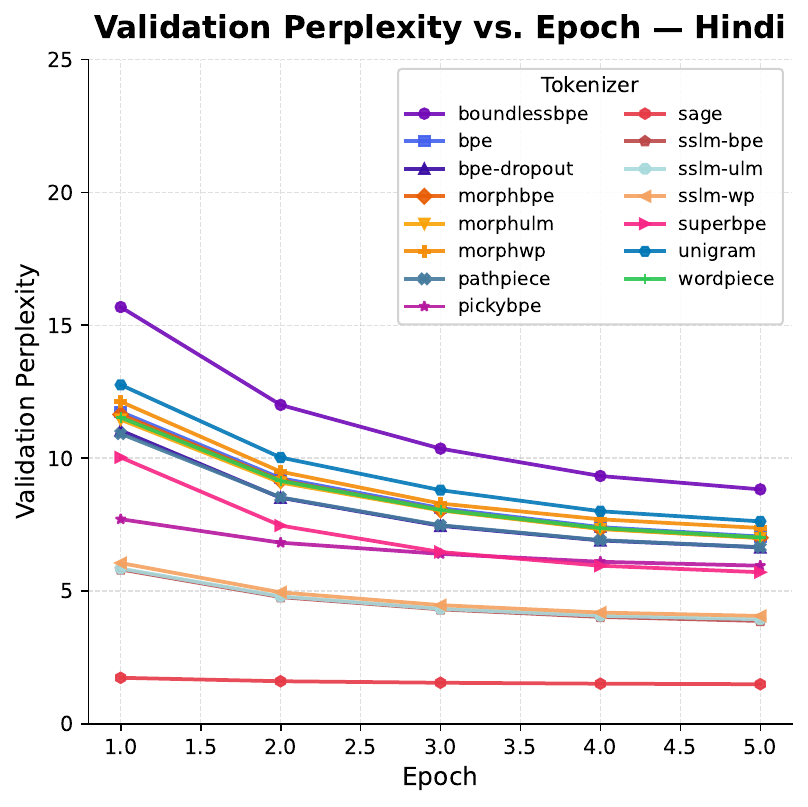}
        \caption{Hindi (Fusional)}
        \label{fig:eval_perplexity_hin}
    \end{subfigure}

    \caption{Variation in the validation perplexity of BERT models during pretraining. Models trained with tokenizers using \g{SSLM} as pretokenization consistently achieve the lowest perplexity.}
    \label{fig:eval_perplexity}
\end{figure*}

\begin{table*}[h]
\centering
\scriptsize
\setlength{\tabcolsep}{4.5pt}
\begin{tabular}{c|
                ccc|cl
                c|ccc|
                cccc}
\toprule
\multirow{1}{*}{\textbf{Tokenizer}} & 

\multicolumn{3}{c}{\textbf{Sentiment Analysis (Acc)}} & 
\multicolumn{3}{c}{\textbf{POS Tagging (F1)}} & 
\multicolumn{3}{c}{\textbf{Named Entity Recognition (F1)}} &
\multicolumn{3}{c}{\textbf{Dependency Parsing (LAS)}}\\
\cmidrule(lr){2-4} \cmidrule(lr){5-7} \cmidrule(lr){8-10} \cmidrule(lr){11-13}
\multicolumn{1}{c}{\textit{Language}} & \textbf{\tiny eng\_latn} & \textbf{\tiny tel\_telu} & \textbf{\tiny hin\_deva}
  & \textbf{\tiny eng\_latn} & \textbf{\tiny tel\_telu} & \textbf{\tiny hin\_deva}
  & \textbf{\tiny eng\_latn} & \textbf{\tiny tel\_telu} & \textbf{\tiny hin\_deva}
  & \textbf{\tiny eng\_latn} & \textbf{\tiny tel\_telu} & \textbf{\tiny hin\_deva} \\
\toprule
\rowcolor[gray]{.95} \multicolumn{13}{l}{\textit{Standard Algorithms}} \\
BPE &
85.55 &
62.85 &
76.67 &
95.09 &
88.41 &
97.15 &
89.98 &
94.72 &
90.69 &
75.64 &
66.31 &
88.20 \\
UnigramLM &
83.61 &
64.88 &
\textbf{77.63} &
95.11 &
89.72 &
97.20 &
89.61 &
94.80 &
91.08 &
75.95 &
68.88 &
\textbf{88.73} \\
WordPiece &
84.86 &
65.91 &
76.10 &
95.22 &
89.51 &
\textbf{97.34} &
88.37 &
\textbf{95.33} &
90.73 &
\textbf{78.11} &
66.31 &
88.63 \\
\midrule
\rowcolor[gray]{.95} \multicolumn{13}{l}{\textit{Contextual Algorithms}} \\
SaGe &
85.32 &
63.22 &
73.61 &
91.68 &
81.99 &
91.67 &
83.32 &
91.88 &
88.32 &
62.17 &
55.14 &
70.07 \\
\midrule
\rowcolor[gray]{.95} \multicolumn{13}{l}{\textit{Inference Modifications}} \\
BPE-dropout &
80.28 &
62.66 &
72.85 &
93.06 &
83.50 &
95.37 &
84.70 &
92.78 &
89.41 &
68.38 &
59.06 &
83.47 \\
PathPiece &
84.75 &
64.32 &
77.06 &
92.27 &
87.10 &
95.08 &
83.89 &
92.62 &
87.04 &
62.76 &
58.76 &
77.47 \\
\midrule
\rowcolor[gray]{.95} \multicolumn{13}{l}{\textit{Vocabulary Modifications}} \\
PickyBPE &
73.16 &
60.44 &
58.70 &
78.74 &
76.57 &
77.54 &
68.96 &
88.61 &
84.27 &
49.96 &
45.77 &
56.66 \\
\midrule
\rowcolor[gray]{.95} \multicolumn{13}{l}{\textit{Cross-token Algorithms}} \\
BoundlessBPE &
83.49 &
65.25 &
75.14 &
91.98 &
83.50 &
95.25 &
84.84 &
93.62 &
87.15 &
61.65 &
57.55 &
78.08 \\
SuperBPE &
83.83 &
64.51 &
74.76 &
93.82 &
85.17 &
96.17 &
88.93 &
93.05 &
88.21 &
68.32 &
62.69 &
80.85 \\
\midrule
\rowcolor[gray]{.95} \multicolumn{13}{l}{\textit{Pre-tokenization using Morfessor}} \\
MorphBPE &
86.24 &
63.96 &
76.10 &
95.18 &
89.12 &
97.24 &
\textbf{90.63} &
94.66 &
91.02 &
75.56 &
\textbf{69.18} &
88.11 \\
MorphULM &
85.89 &
66.91 &
77.06 &
\textbf{95.28} &
89.38 &
97.33 &
90.03 &
94.90 &
90.52 &
75.28 &
69.03 &
88.24 \\
MorphWPC &
\textbf{87.38} &
\textbf{67.28} &
75.91 &
95.26 &
90.03 &
97.23 &
88.10 &
94.95 &
90.75 &
77.65 &
69.03 &
88.35 \\
\midrule
\rowcolor[gray]{.95} \multicolumn{13}{l}{\textit{Pre-tokenization using SSLM}} \\
SSLM-BPE &
\underline{85.66}\textsuperscript{\textcircled{4}} &
\underline{64.70}\textsuperscript{\textcircled{7}} &
\underline{77.06}\textsuperscript{\textcircled{4}} &
95.09\textsuperscript{\textcircled{6}} &
\underline{88.42}\textsuperscript{\textcircled{8}} &
\underline{97.18}\textsuperscript{\textcircled{6}} &
88.93\textsuperscript{\textcircled{5}} &
94.38\textsuperscript{\textcircled{9}} &
\underline{91.03}\textsuperscript{\textcircled{4}} &
74.90\textsuperscript{\textcircled{8}} &
\underline{66.77}\textsuperscript{\textcircled{7}} &
87.47\textsuperscript{\textcircled{8}} \\
SSLM-ULM &
\underline{84.17}\textsuperscript{\textcircled{9}} &
63.40\textsuperscript{\textcircled{11}} &
76.10\textsuperscript{\textcircled{8}} &
95.06\textsuperscript{\textcircled{8}} &
\underline{\textbf{90.17}}\textsuperscript{\textcircled{1}} &
97.16\textsuperscript{\textcircled{8}} &
88.88\textsuperscript{\textcircled{6}} &
94.58\textsuperscript{\textcircled{8}} &
\underline{91.30}\textsuperscript{\textcircled{2}} &
73.35\textsuperscript{\textcircled{9}} &
68.58\textsuperscript{\textcircled{6}} &
86.95\textsuperscript{\textcircled{9}} \\
SSLM-WPC &
84.75\textsuperscript{\textcircled{10}} &
65.25\textsuperscript{\textcircled{5}} &
74.38\textsuperscript{\textcircled{12}} &
94.95\textsuperscript{\textcircled{9}} &
\underline{90.10}\textsuperscript{\textcircled{2}} &
97.17\textsuperscript{\textcircled{7}} &
87.05\textsuperscript{\textcircled{9}} &
94.84\textsuperscript{\textcircled{6}} &
\underline{\textbf{91.41}}\textsuperscript{\textcircled{1}} &
76.78\textsuperscript{\textcircled{3}} &
\underline{68.73}\textsuperscript{\textcircled{5}} &
87.51\textsuperscript{\textcircled{7}} \\
\bottomrule
\end{tabular}
\caption{Downstream performance of 12M-parameter BERT models trained with different tokenizer variants. Best results for each task-language pair are in \textbf{bold}; superscripts denote tokenizer ranks with SSLM pretokenization. Scores are averaged over three finetuning runs (seeds 42, 43, 44). We report accuracy for Sentiment Analysis, F1 for POS tagging and NER, and LAS for Dependency Parsing. Variants improved by \g{SSLM} pretokenization are \underline{underlined}.}
\label{tab:downstream_performance}
\end{table*}

We measure pairwise Jaccard overlap on the test split to compare tokens learned by \textit{tokenizer-free} and \textit{fixed-tokenizer} approaches across language typologies. Figure \ref{fig:vocab_overlap} shows the corresponding heatmap.

\textbf{Findings:} \bl{H-Net}s exhibit near-zero overlap with fixed tokenizers across typologies, indicating that end-to-end data-dependent boundary prediction learns token sets fundamentally different from frequency-based subword methods by prioritizing byte-level sequence efficiency over standard subword boundaries. \g{SSLMs}, on the other hand, show comparatively higher (though still low) overlap with fixed tokenizers, particularly with frequency-based methods such as \ora{BPE} and \ora{WPC}; the strongest similarity occurs between \g{SSLM}s and \ora{BPE} variants across all typologies.
Overlap is lowest for agglutinative languages ($\approx0.12$ to $0.18$ Jaccard), suggesting that productive morphology generates many possible subwords, encouraging joint optimization to learn smaller atomic morphemes while fixed tokenizers merge them into larger but linguistically arbitrary units.
\ora{BPE}, \ora{BPE-dropout}, \ora{PathPiece}, and \ora{MorphBPE} form a high-overlap cluster (greater than $0.80$ Jaccard), indicating that changes to pre-tokenization or inference do not fundamentally alter learned vocabularies. In contrast, \ora{SaGe} shows lower overlap, suggesting that prioritizing contextual exponence substantially changes the vocabulary.

\subsection{Q4: Are the learned tokens better?}
\label{sec:downstream_evaluation}

To determine if these fundamentally distinct tokens learned through \textit{tokenizer-free} approaches improve language understanding, we analyze the downstream performance of BERT \cite{devlin-etal-2019-bert} models pretrained with each tokenizer and finetuned for the tasks listed in \S\ \ref{sec:evaluation}, with results reported in Table \ref{tab:downstream_performance}. As noted, \textit{tokenizer-free} approaches do not have a fixed vocabulary, while BERT models require static, fixed, pre-defined vocabulary. Therefore, we use \textit{tokenizer-free} approaches\footnote{We avoid considering pretokenization with \bl{H-Net}s, as it is observed that it mostly segments at whitespaces and it is not straightforward to limit and fix its vocabulary due to its cross-token flexibility.} as pretokenization and learn a standard fixed tokenizer on the pre-segmented corpus. This approach fixes the model architecture and provides a fair downstream comparison. We refer to these with prefix \g{SSLM}, i.e., \g{SSLM-BPE}, \g{SSLM-ULM}, and \g{SSLM-WPC}.

\textbf{Findings:} As illustrated in Figure \ref{fig:eval_perplexity} and summarized in Table \ref{tab:perplexity_table}, \textit{tokenizer-free} approaches consistently achieve the lowest validation perplexity across all three evaluated languages and converge earlier compared to fixed tokenizers. Only in Hindi, we observe \ora{SaGe} outperforming \textit{tokenizer-free} approaches with nearly perfect perplexity of $1$. This improvement in perplexity is even more pronounced as compared to other pretokenization-based modifications such as that of \ora{Morfessor}. We also observe that Telugu has the highest perplexity without pretokenization, but the lowest perplexity when the \textit{tokenizer-free} \g{SSLM} is used for pretokenization.

\begin{table}[H]
\centering
\scriptsize
\begin{tabular}{lccc}
\toprule
\textbf{Pretokenizer} & \textbf{eng\_latn} & \textbf{tel\_telu} & \textbf{hin\_deva} \\
\midrule
None & 9.84 & 9.92 & 7.23 \\

SSLM 
& \textbf{3.81} {\scriptsize($\downarrow$61.3\%)} 
& \textbf{2.92} {\scriptsize($\downarrow$70.6\%)} 
& \textbf{3.96} {\scriptsize($\downarrow$45.2\%)} \\

\bottomrule
\end{tabular}
\caption{Aligning tokens learned by fixed-tokenizer approaches with that learned by \g{SSLM} through pretokenization significantly reduces the perplexity, showcasing improvement in language modeling. The scores are averages of perplexities across \ora{BPE}, \ora{ULM}, and \ora{WPC}.}
\label{tab:perplexity_table}
\end{table}

The downstream results (see Table \ref{tab:downstream_performance} above) showcase the practical implications of each approach. Despite having fundamentally distinct vocabulary, \textit{tokenizer-free} approaches remain competitive. In contrast, the performance of many modified \textit{fixed-tokenizer} approaches such as \ora{SaGe}, \ora{PickyBPE}, and \ora{BPE-dropout} remains inconsistent and often significantly lower across different tasks and languages. Notably, \g{SSLM-ULM} achieves the highest F1-score for POS tagging in Telugu ($90.17\%$), while \g{SSLM-WPC} achieves highest for NER in Hindi ($91.41\%$). Pretokenization with \ora{Morfessor} remains the best performing approach overall. It consistently improves the performance over standard approaches. Overall, the results suggest that jointly learned tokenizers learn meaningful tokens that transfer effectively to downstream NLP tasks.

We additionally perform an ablation study across different model scales and observe consistent trends. \g{SSLM}-based pretokenization continues to achieve lower validation perplexity and competitive downstream performance across scales, although the relative gains reduce slightly for larger models. Detailed experimental setup, hyperparameters, and complete results are provided in the \S\ \ref{sec:ablation-model-scale} of Appendix.

\section{Conclusion}

In this work, we showed that jointly learned tokenization fundamentally changes the structure and behavior of learned tokens across typologically diverse languages. In particular, \g{SSLM}s learn more morphologically aligned and contextually efficient segments than standard frequency-based tokenizers, while \bl{H-Net}s prioritize byte-level efficiency and produce substantially different token structures. Although morphology-aware pretokenization remains strongest overall, \g{SSLM}-based pretokenization consistently remains competitive on downstream performance. Overall, our findings show that jointly optimized tokenizers can learn linguistically meaningful token structures that remain effective for downstream NLP despite having substantially different vocabularies from standard tokenization approaches.\\ 

\pagebreak

\section*{Limitations}

While we do our best to handle confounding factors such as dataset sizes, model sizes, and other experimental setup, few limitations remain:
\begin{enumerate}[leftmargin=*, labelindent=0pt, itemindent=0pt]
  \setlength\itemsep{0em}
  \item \textbf{Model and dataset sizes}: We focused on spanning an extensive set of tokenization approaches, languages, and their combinations. Therefore, our analysis was limited to models and dataset at relatively smaller scale. For instance, we limit \textit{tokenizer-free} LMs to 3M parameters and dataset of 250,000 sentences. Downstream evaluation was performed on BERT models at 12M parameters and dataset of 10M sentences. Although we perform an ablation study with \ora{BPE} and \g{SSLM-BPE} tokenization approaches, we leave scalability analysis of our findings as future work. In this work, we trade-off model and dataset sizes for larger coverage of languages and tokenization approaches.
  \item \textbf{Generalizability of downstream results}: For every combination of tokenization approach and language, we had to pretrain a model. Hence, our downstream evaluation was limited to three typologically distinct languages: English, Hindi, and Telugu. We do not evaluate downstream performance across all 18 languages, in order to keep computational cost manageable. We focus on evaluating all tokenization approaches for these selected languages in a controlled experimental setting. 
  \item \textbf{Hyperparameters of \textit{tokenizer-free} LMs}: We do not sweep over certain hyperparameters of \textit{tokenizer-free} LMs, for example, maximum token length in \g{SSLM}. We fix them following heuristics from previous works, since sweeping over these hyperparameters require significant computation resources, which we limit.
\end{enumerate}



\section*{Ethics Statement}

This work focuses on the analysis of tokenization strategies across languages and does not involve human subjects or the use of sensitive personal data. All experiments are conducted on publicly available text corpora. However, differences in dataset quality, and linguistic coverage across languages may introduce biases that affect the observed behavior of tokenization methods. 


\bibliography{custom}

\clearpage

\appendix

\label{sec:appendix}

\section{Ablation Study: Model Scale}
\label{sec:ablation-model-scale}
In our downstream evaluation, we fix the size of BERT models at approximately 12M parameters. This was scaled accordingly with respect to the data size of 10M sentences. Here, we perform an ablation study by changing the size of our models while keeping the data size same. We perform additional evaluation of BERT models at approximately 2M and 30M parameters. To limit the computational resources, we limit our analysis to two tokenization approaches: \ora{BPE} and \g{SSLM-BPE}. This allows us to focus on the affect of pretokenization using \textit{tokenizer-free} approach \g{SSLM}.

\begin{table*}[h]
\centering
\scriptsize
\setlength{\tabcolsep}{4.5pt}
\begin{tabular}{c|
                ccc|cl
                c|ccc|
                cccc}
\toprule
\multirow{1}{*}{\textbf{Model Scale}} & 

\multicolumn{3}{c}{\textbf{Sentiment Analysis (Acc)}} & 
\multicolumn{3}{c}{\textbf{POS Tagging (F1)}} & 
\multicolumn{3}{c}{\textbf{Named Entity Recognition (F1)}} &
\multicolumn{3}{c}{\textbf{Dependency Parsing (LAS)}}\\
\cmidrule(lr){2-4} \cmidrule(lr){5-7} \cmidrule(lr){8-10} \cmidrule(lr){11-13}
\multicolumn{1}{c}{\textit{Language}} & \textbf{\tiny eng\_latn} & \textbf{\tiny tel\_telu} & \textbf{\tiny hin\_deva}
  & \textbf{\tiny eng\_latn} & \textbf{\tiny tel\_telu} & \textbf{\tiny hin\_deva}
  & \textbf{\tiny eng\_latn} & \textbf{\tiny tel\_telu} & \textbf{\tiny hin\_deva}
  & \textbf{\tiny eng\_latn} & \textbf{\tiny tel\_telu} & \textbf{\tiny hin\_deva} \\
\toprule
\rowcolor[gray]{.95} \multicolumn{13}{l}{\textit{Standard Algorithm: BPE}} \\
2M &
77.75 &
59.15 &
65.97 &
89.16 &
62.22 &
92.57 &
70.79 &
86.15 & 
75.90 &
54.78 &
56.04 &
74.12 \\
12M &
85.55 &
62.85 &
76.67 &
95.09 &
88.41 &
97.15 &
89.98 &
94.72 &
90.69 &
75.64 &
66.31 &
88.20 \\
30M &
\textbf{88.42} &
\textbf{65.43} &
\textbf{78.01} &
95.97 &
\textbf{92.82} &
\textbf{97.64} &
\textbf{91.65} &
\textbf{95.66} &
91.91 &
\textbf{80.29} &
\textbf{72.96} &
\textbf{90.44} \\
\midrule
\rowcolor[gray]{.95} \multicolumn{13}{l}{\textit{Using SSLM as pretokenization: SSLM-BPE}} \\
2M &
76.95 &
\underline{59.89} &
58.89 &
88.05 &
61.76 &
91.28 &
56.18 &
83.37 &
\underline{76.61} &
50.99 &
53.62 &
72.38 \\
12M &
\underline{85.66} &
\underline{64.70} &
\underline{77.06} &
95.09 &
\underline{88.42} &
\underline{97.18} &
88.93 &
94.38 &
\underline{91.03} &
74.90 &
\underline{66.77} &
87.47 \\
30M &
86.01 &
64.70 &
78.01 &
\underline{\textbf{96.13}} &
92.78 &
97.45 &
91.59 &
\underline{\textbf{95.70}} &
\underline{\textbf{92.72}} &
80.20 &
70.09 &
89.88 \\
\bottomrule
\end{tabular}
\caption{Downstream performance of BERT models at different scales trained with tokenizers BPE and SSLM-BPE. Best results for each task-language pair are in \textbf{bold}. Scores are averaged over three finetuning runs (seeds 42, 43, 44). We report accuracy for Sentiment Analysis, F1 for POS tagging and NER, and LAS for Dependency Parsing. Variants improved by \g{SSLM} pretokenization are \underline{underlined}.}
\label{tab:downstream_performance_model_scales}
\end{table*}
\begin{figure}[h]
  \centering
  \includegraphics[width=\columnwidth]{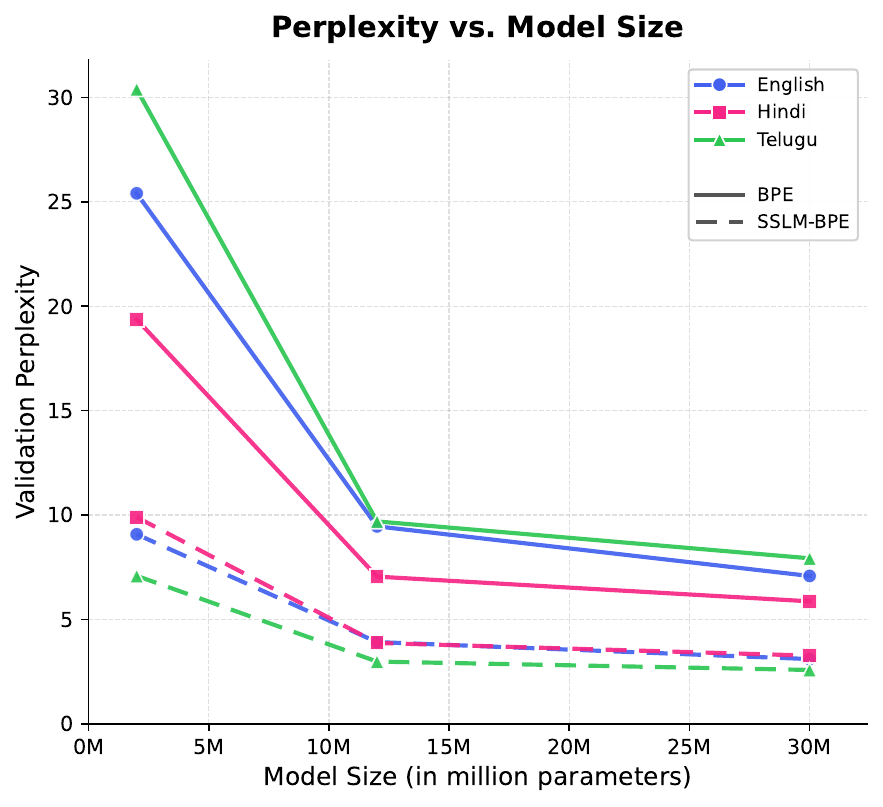}
  \caption{Validation perplexity of BERT models at different scales. Aligning tokens with \textit{tokenizer-free} approaches through pretokenization consistently results in lower perplexity.}
  \label{fig:model-size-perplexity}
\end{figure}
\begin{figure*}[t]
  \centering
  \begin{subfigure}[t]{0.48\textwidth}
    \centering
    \includegraphics[width=\linewidth]{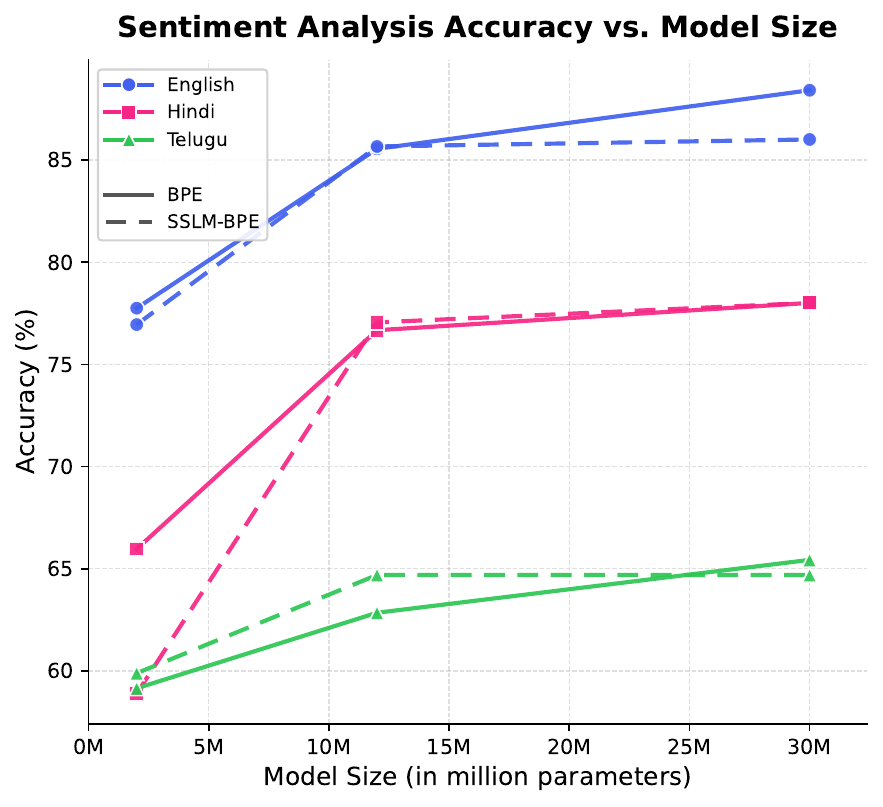}
    \caption{Sentiment analysis (Accuracy)}
    \label{fig:model-scale-sentiment}
  \end{subfigure}
  \hfill
  \begin{subfigure}[t]{0.48\textwidth}
    \centering
    \includegraphics[width=\linewidth]{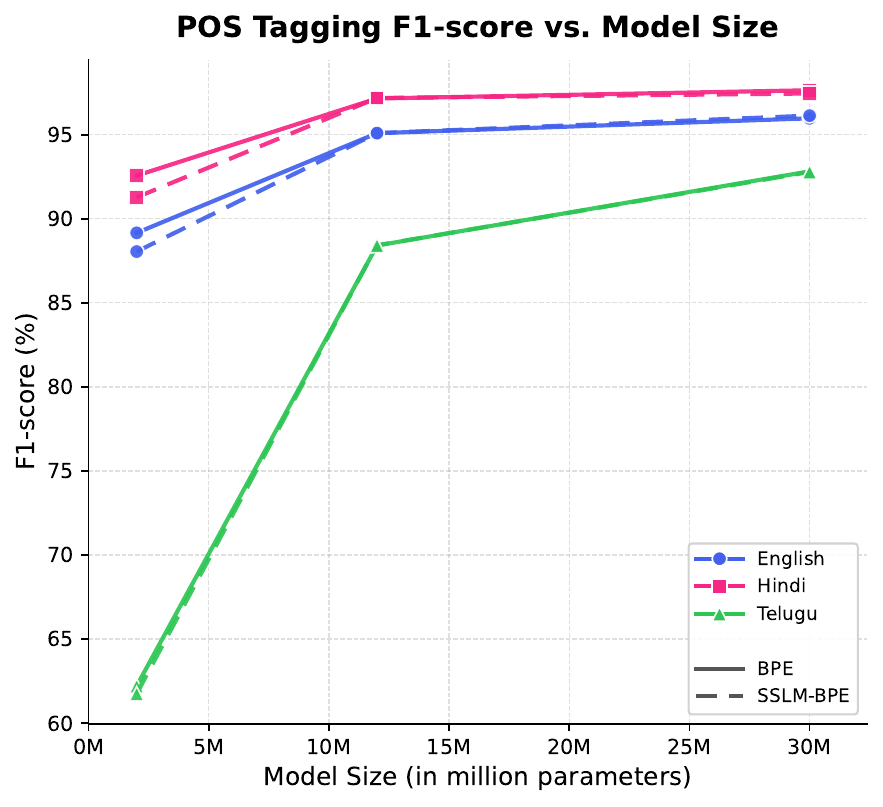}
    \caption{POS tagging (F1-score)}
    \label{fig:model-scale-pos}
  \end{subfigure}


  \begin{subfigure}[t]{0.48\textwidth}
    \centering
    \includegraphics[width=\linewidth]{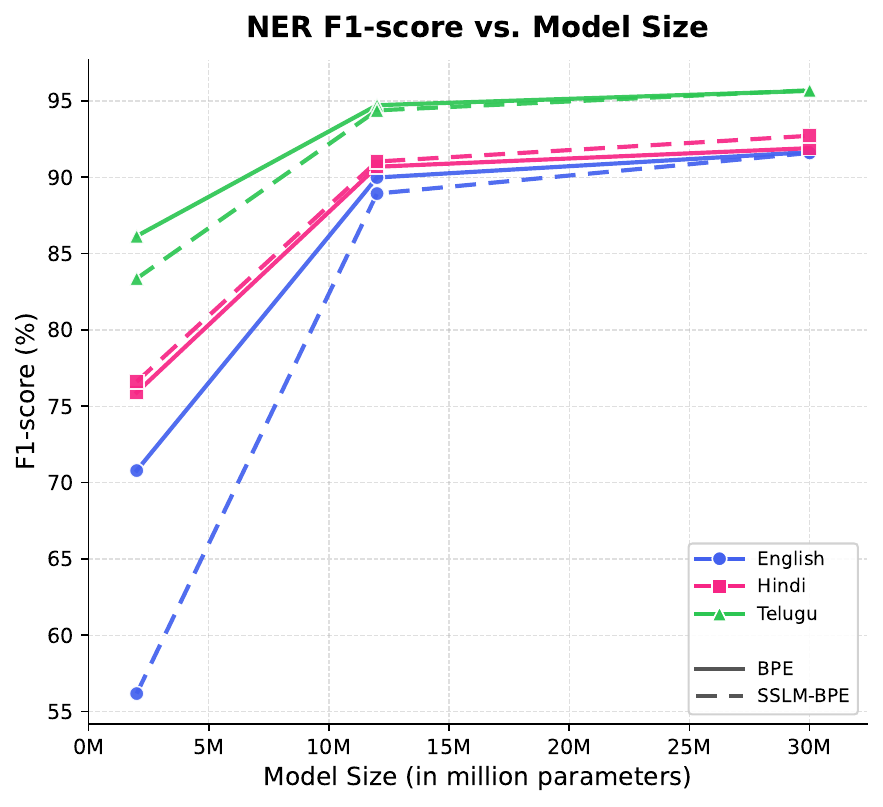}
    \caption{NER (F1-score)}
    \label{fig:model-scale-ner}
  \end{subfigure}
  \hfill
  \begin{subfigure}[t]{0.48\textwidth}
    \centering
    \includegraphics[width=\linewidth]{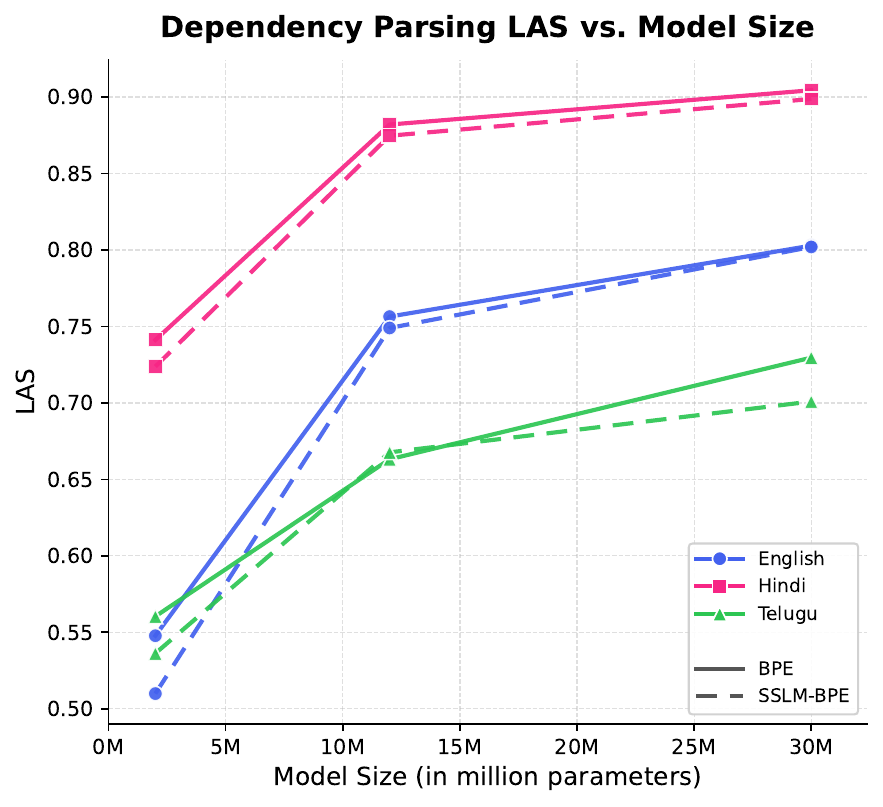}
    \caption{Dependency parsing (LAS)}
    \label{fig:model-scale-dep}
  \end{subfigure}
  \caption{Performance of BERT models across model scales on sentiment analysis, POS tagging, NER, and dependency parsing.}
  \label{fig:model-size-performance}
\end{figure*}
\textbf{Findings}: As observed in Figure \ref{fig:model-size-perplexity}, models trained using \g{SSLM} as pretokenization consistently show lower perplexity across all model sizes. However, as model size increases, the improvement saturates. This indicates that \g{SSLM-BPE} generally provides more efficient language modeling. 
We also perform evaluation on downstream tasks at different scales. The results are listed in Table \ref{tab:downstream_performance_model_scales} and plotted in Figure \ref{fig:model-size-performance}. Across all three languages, \g{SSLM-BPE} often matches \ora{BPE} on downstream tasks such as POS tagging and NER. We observe that whenever model benefits with larger size, the gap between \g{SSLM-BPE} and \ora{BPE} reduces, especially for POS tagging and NER tasks.
\textbf{Note}: Our \textit{tokenizer-free} LMs were trained 250,000 sentence corpora, rather than on the entire 10M sentence corpora which we use for downstream evaluation. This was due to high computation requirement of such models (approximately $10\times$ slower than transformers). We leave further scaling analysis accounting for these factors as our future work.

\section{Hyperparameters \& Experimental Setup}
\label{sec:hyperparameters}

\subsection{H-Nets}
\label{sec:h-net-experimental-setup}

Table \ref{tab:hnet_hyperparameters} details the primary hyperparameter configurations used for pretraining the Hierarchical Network (\bl{H-Net}) models across all evaluated languages. The model employs a structural layout of $m_1$ at the bottom level and $T_2$ at the top level, utilizing an embedding and hidden dimension of 256 for both levels. The bottom level ($m_1$) operates purely without an FFN block, while the top level ($T_2$) uses an intermediate feed-forward dimension of 640. This setup operates directly on byte sequences (vocabulary size of 256) and results in a lightweight model with approximately 3M parameters. We trained the model using an AdamW optimizer (with $\beta_1=0.9, \beta_2=0.95$) and a batch size of 128 for a maximum of 70 epochs. Optimization was guided by a cosine learning rate scheduler with a peak base learning rate of $3 \times 10^{-4}$ after an initial 10\% linear warmup phase. To prevent overfitting, early stopping was employed with a patience of 5 epochs alongside a weight decay of 0.01. The \bl{H-Net} specific structural compression constraints were maintained constantly throughout training, with the ratio loss scale set to 1.0 and zero warmup compression epochs.

\begin{table}[h]
\centering
\small 
\begin{tabular}{@{}lc@{}}
\toprule
\textbf{Hyperparameter} & \textbf{Value} \\
\midrule
\multicolumn{2}{c}{\textit{Model Architecture}} \\
\midrule
Architecture Type & H-Net \\
Total Parameters & $\sim 3M$ \\
Architecture Layout & $m_1 T_2$ \\
Embedding Dimension & $256$ \\
Feed-Forward Dimension & $640$ \\
Attention Heads & $4$ \\
\midrule
\multicolumn{2}{c}{\textit{Optimization Setup}} \\
\midrule
Optimizer & AdamW \\
Adam $\beta_1, \beta_2$ & $(0.9, 0.95)$ \\
Learning Rate & $0.0003$ \\
Learning Rate Scheduler & Cosine \\
Warmup Ratio & $10\%$ of total steps \\
Weight Decay & $0.01$ \\
Max Epochs & $70$ \\
Early Stopping Patience & $5$ \\
\midrule
\multicolumn{2}{c}{\textit{H-Net \& Data Configuration}} \\
\midrule
Vocabulary Size (Bytes) & $256$ \\
Max Sequence Length & $4096$ \\
Batch Size & $128$ \\
Ratio Loss Scale & $1.0$ \\
Warmup Compression Epochs & $0$ \\
\bottomrule
\end{tabular}
\caption{Hyperparameter configurations and approximate parameter count for the Hierarchical Network (\bl{H-Net}) pre-training phase.}
\label{tab:hnet_hyperparameters}
\end{table}

\subsection{SSLMs}
\label{sec:sslm-experimental-setup}

Table~\ref{tab:sslm_hyperparameters} details the comprehensive set of hyperparameters utilized during the pre-training of our Subword Segmental Language Models (\g{SSLM}s) across all evaluated languages. We adopted a lightweight Transformer decoder architecture tailored to subword segmental modeling, bringing the total parameter count to 2,105,836. Key architectural adjustments include scaling down the number of layers, embedding dimensions, and feed-forward dimensions to ensure fast and efficient training with a limited parameter budget. The maximum segment length was constrained to 5 spatial units to balance compositional flexibility and computational overhead. Optimization was performed using the Adam optimizer coupled with an inverse square root learning rate scheduler and a linear warmup phase.
\begin{table}[h]
\centering
\small 
\begin{tabular}{@{}lc@{}}
\toprule
\textbf{Hyperparameter} & \textbf{Value} \\
\midrule
\multicolumn{2}{c}{\textit{Model Architecture}} \\
\midrule
Architecture Type & Transformer SSLM \\
Total Parameters & $\sim 2M$ \\
Decoder Layers & $3$ \\
Embedding Dimension & $128$ \\
Feed-Forward Dimension & $512$ \\
Attention Heads & $4$ \\
\midrule
\multicolumn{2}{c}{\textit{Optimization Setup}} \\
\midrule
Optimizer & Adam \\
Adam $\beta_1, \beta_2$ & $(0.9, 0.98)$ \\
Learning Rate & $0.0005$ \\
Learning Rate Scheduler & Inverse Square Root \\
Warmup Updates & $1000$ \\
Initial Warmup Learning Rate & $1\times 10^{-7}$ \\
Weight Decay & $0.01$ \\
Gradient Clipping Norm & $0.0$ \\
Dropout & $0.1$ \\
Max Epochs & $25$ \\
Early Stopping Patience & $2$ \\
\midrule
\multicolumn{2}{c}{\textit{SSLM \& Data Configuration}} \\
\midrule
Max Segment Length & $5$ \\
Lexicon Max Size & $10000$ \\
Tokens Per Sample & $512$ \\
Max Tokens & $16384$ \\
\bottomrule
\end{tabular}
\caption{Hyperparameter configurations and parameter count for the lightweight Subword Segmental Language Model (SSLM) pre-training phase.}
\label{tab:sslm_hyperparameters}
\end{table}

\subsection{Fixed-tokenizer Variants}

\label{subsec:tokenizer_hyperparameters}

To ensure a fair and comprehensive comparison across different tokenization strategies, we standardized the training hyperparameter configurations where applicable, while setting strategy-specific parameters to their best-known or recommended values.

\paragraph{Global Settings} 
All tokenizers were trained across three final target vocabulary sizes: $T \in \{5000, 10000, 20000\}$. For tokenizers relying on the SentencePiece library (e.g., standard \ora{BPE}, \ora{Unigram}, and \ora{PathPiece}), the character coverage parameter was explicitly set to $1.0$ (100\%) to prevent any heuristic character dropping and uniformly ensure complete representation of the datasets' alphabets. We consider fixed-tokenizer variants with final vocabulary size of $10,000$ for a fair comparison with \g{SSLM}s with initial lexicon of size $10,000$.

\paragraph{Standard Subword Architectures} 
For \textbf{WordPiece}, we utilized the HuggingFace tokenizers library, configuring it with NFKC Unicode normalization and lowercasing. Pre-tokenization was handled via a standard whitespace split, and the continuation prefix was assigned as ``\texttt{\#\#}''. The standard special tokens strictly enforced were \texttt{[UNK], [CLS], [SEP], [PAD]}, and \texttt{[MASK]}. Standard \textbf{\ora{BPE}} and \textbf{\ora{UnigramLM}} models relied on SentencePiece defaults beyond the mandatory complete character coverage.

\paragraph{SuperBPE}
Following the two-stage subword acquisition strategy of \ora{SuperBPE}, the base vocabulary size (the transition limit $t$ between stage 1 and stage 2) was configured at $t = 4000$ when final vocabulary size desired was $T = 10000$, and $t = 10000$ when $T = 25000$ (i.e., $T = t \times2.5$, as found in \citet{liu2025superbpespacetravellanguage}).

\paragraph{BoundlessBPE} 
For \ora{BoundlessBPE}, the $\tau$ parameter, managing the trade-off threshold, was set to $0.9$. The vocabulary pruning recalculation interval was configured to occur every $1000$ iterations (\texttt{recalc = 1000}). We utilized the \texttt{ultimate2} regular expression split pattern constraints, taking advantage of the active vocabulary blowup tolerance (\texttt{blowup = 1}). 

\paragraph{PickyBPE}
The intersection-over-sequence threshold regulating token acceptance in \ora{PickyBPE} was strictly confined to $0.9$.

\paragraph{Morphologically-Informed Tokenizers (MorphBPE, MorphULM, MorphWP)}
To create tokenizers aligned with linguistic boundaries, an unsupervised \textbf{\ora{Morfessor}} Baseline model was first trained on the raw corpus to dictate morphological decisions. Texts were pre-tokenized along the predicted morphological seams, creating a pseudo-corpus separated by whitespace. The subsequent \ora{BPE}, \ora{Unigram}, and \ora{WordPiece} counterparts (\textbf{\ora{MorphBPE}}, \textbf{\ora{MorphULM}}, and \textbf{\ora{MorphWP}}) were exclusively trained on this constrained configuration, forcing the statistically induced vocabularies to respect Morfessor's proposed boundaries.

\subsection{Downstream Evaluation}
\label{sec:downstream-evaluation-exp-setup}

As discussed, we perform downstream evaluation by pretraining BERT \cite{devlin-etal-2019-bert} models and later finetuning them on each task. Table \ref{tab:bert_hyperparameters} lists the hyperparameters we used while pretraining BERT models. Similarly, Table \ref{tab:bert_finetuning_hyperparams} lists the hyperparameters while finetuning for each task.

We describe our downstream tasks as follows: \\
\textbf{POS tagging}: Involves assigning a grammatical category such as noun, verb, adjective, to each word in a sentence. It involves reasoning through both the definition of the word and its context. We consider the class of word as the class corresponding to its first subword. We use Universal Dependency (UD) treebanks \cite{nivre-etal-2020-universal} via HuggingFace universal dependency dataset: \texttt{en\_ewt} for English, \texttt{te\_mtg} for Telugu, and \texttt{hi\_hdtb} for Hindi. Labels are drawn from the 17 universal POS tags (\textsc{upos}). Performance is measured with weighted-average F1-score.\\
\textbf{Sentiment Analysis}: This tasks involves determining the sentiment of a sentence, e.g., positive, negative, or neutral. We use ACTSA (Annotated Corpus for Telugu Sentiment Analysis) dataset for Telugu \cite{mukku-mamidi-2017-actsa}, IITP Product Review dataset \cite{akhtar-etal-2016-hybrid}, and SST-2 dataset from GLUE benchmark \cite{wang-etal-2018-glue}. \\ 
\textbf{Named Entity Recognition} (NER): This tasks involves identifying and classifying named entities into predefined categories such as persons, organizations, locations, etc. We use CoNLL NER dataset \cite{tjong-kim-sang-de-meulder-2003-introduction} for English, and WikiAnn dataset \cite{pan-etal-2017-cross, doddapaneni-etal-2023-towards}. WikiAnn consists of coarse grained labels as follows: Person (POS), Organization (ORG), and Location (LOC). Whereas, CoNLL NER contains an additional Miscellaneous (MISC) tag.\\ 
\textbf{Dependency Parsing} (DP): This task involves analyzing the grammatical structure of a sentence by identifying the relationship between head words and their dependents. We again use Universal Dependency (UD) treebank dataset \cite{nivre-etal-2020-universal}. We report Labeled Attachment Score (LAS) \cite{nivre-fang-2017-universal}.

\begin{table}[t]
\centering
\small
\begin{tabular}{@{}lc@{}}
\toprule
\textbf{Hyperparameter} & \textbf{Value} \\
\midrule
\multicolumn{2}{c}{\textit{Model Architecture}} \\
\midrule
Architecture Type       & BERT \\
Total Parameters        & $\sim$2M / $\sim$12M / $\sim$30M \\
Hidden Layers           & 3 / 6 / 8 \\
Hidden Dimension        & 128 / 384 / 512 \\
Feed-Forward Dimension  & 512 / 1024 / 2048 \\
Attention Heads         & 4 / 6 / 8 \\
Max Position Embeddings & 128 \\
Activation Function     & \texttt{gelu} \\
\midrule
\multicolumn{2}{c}{\textit{Regularization}} \\
\midrule
Attention Dropout       & $0.1$ \\
Hidden Dropout          & $0.1$ \\
Initializer Range       & $0.01$ / $0.02$ / $0.02$ \\
Layer Norm $\epsilon$   & $1 \times 10^{-12}$ \\
\midrule
\multicolumn{2}{c}{\textit{Pre-Training Configuration}} \\
\midrule
Pre-Training Objective  & Masked LM \\
MLM Masking Probability & $0.15$ \\
Sequence Length         & $128$ \\
Max Epochs              & $5$ \\ 
Early Stopping Patience & $1$ \\
\bottomrule
\end{tabular}
\caption{Architecture and pre-training configurations for the three BERT model scales (2M / 12M / 30M). Where values differ across scales, they are listed in order from smallest to largest model.}
\label{tab:bert_hyperparameters}
\end{table}

\begin{table}[t]
\small
\centering
\begin{tabular}{@{}lp{3.6cm}@{}}
\toprule
\textbf{Hyperparameter} & \textbf{Value} \\
\midrule
Train batch size   & 32 \\
Eval batch size    & 32 \\
Epochs             & 3 for GLUE \\
                   & 10 for ACTSA-TE, Dep.\ Parsing,
                     IITP-PR, NER (CoNLL), POS,
                     Wiki-NER \\
Learning rate      & 2e-5 \\
LR schedule        & Linear warmup \\
Warmup ratio       & 10\% of steps \\
Weight decay       & 0.01 \\
Adam $\epsilon$    & 1e-8 \\
Adam $\beta_1$     & 0.9 \\
Adam $\beta_2$     & 0.999 \\
Max sequence length & 128 \\
Precision          & FP16 \\
\bottomrule
\end{tabular}
\caption{Hyperparameters used for fine-tuning BERT across all downstream tasks.}
\label{tab:bert_finetuning_hyperparams}
\end{table}


\begin{table*}[t]
\centering
\fontsize{6.5}{9}\selectfont   
\setlength{\tabcolsep}{4pt}
\setlength{\tabcolsep}{4pt}
\begin{tabular}{ccc ccc ccc ccc ccc ccc}
\toprule

\multicolumn{3}{c}{\textbf{English}} &
\multicolumn{3}{c}{\textbf{Hebrew}} &
\multicolumn{3}{c}{\textbf{Hindi}} &
\multicolumn{3}{c}{\textbf{Tamil}} &
\multicolumn{3}{c}{\textbf{Hungarian}} &
\multicolumn{3}{c}{\textbf{Indonesian}} \\
\midrule
\multicolumn{3}{c}{Bangladeshis} &
\multicolumn{3}{c}{\htok{ריכוזיים}} &
\multicolumn{3}{c}{\ditok{डिस्ट्रीब्यूटरों}} &
\multicolumn{3}{c}{\tatok{தொடுக்க}} &
\multicolumn{3}{c}{zeneszerzőzseniket} &
\multicolumn{3}{c}{kehabisan} \\

\multicolumn{3}{c}{Bangladeshi s} &
\multicolumn{3}{c}{\htok{ריכוזי ים}} &
\multicolumn{3}{c}{\ditok{डिस्ट्रीब्यूटर ों}} &
\multicolumn{3}{c}{\tatok{தொடு க்க}} &
\multicolumn{3}{c}{zeneszerzőzseni ket} &
\multicolumn{3}{c}{ke habis an} \\

\midrule

\multicolumn{2}{c}{\textit{Segmentation}} & \textit{F1} &
\multicolumn{2}{c}{\textit{Segmentation}} & \textit{F1} &
\multicolumn{2}{c}{\textit{Segmentation}} & \textit{F1} &
\multicolumn{2}{c}{\textit{Segmentation}} & \textit{F1} &
\multicolumn{2}{c}{\textit{Segmentation}} & \textit{F1} &
\multicolumn{2}{c}{\textit{Segmentation}} & \textit{F1} \\

\midrule

\multicolumn{2}{c}{\seg{0.0}{Ba ngl ades his}}          & 0   &
\multicolumn{2}{c}{\seg{0.0}{\htok{ריכו ז יים}}}  & 0   &
\multicolumn{2}{c}{\seg{0.0}{\ditok{डिस् ट्री ब्यू टरों}}}  & 0   &
\multicolumn{2}{c}{\seg{0.0}{\tatok{தொ டுக்க}}}       & 0   &
\multicolumn{2}{c}{\seg{0.0}{zene szerz őz sen iket}}    & 0   &
\multicolumn{2}{c}{\seg{0.0}{keha bisa n}}        & 0   \\

\multicolumn{2}{c}{\seg{0.5}{Ba ngla deshi s}}         & 0.5 &
\multicolumn{2}{c}{\seg{0.67}{\htok{ריכ וזי ים}}} & 0.67  &
\multicolumn{2}{c}{\seg{0.33}{\ditok{डिस् ट्री ब्यू ट र ों}}}  & 0.33 &
\multicolumn{2}{c}{\seg{0.67}{\tatok{தொட ு க்க}}}         & 0.67   &
\multicolumn{2}{c}{\seg{0.29}{zene szerz ő zs en i ket}}     & 0.29  &
\multicolumn{2}{c}{\seg{0.80}{ke habi s an}}       & 0.8 \\

\multicolumn{2}{c}{\seg{0.5}{Ba ngla deshi s}}           & 0.5    &
\multicolumn{2}{c}{\seg{0.5}{\htok{רי כוז י ים}}} & 0.5 &
\multicolumn{2}{c}{\seg{0.33}{\ditok{डिस् ट्री ब्यू ट र ों}}}  & 0.33 &
\multicolumn{2}{c}{\seg{0.0}{\tatok{தொ டுக்க}}}         & 0    &
\multicolumn{2}{c}{\seg{0.0}{zene szerz ő zs en ik et}}     & 0    &
\multicolumn{2}{c}{\seg{0.5}{keha bis an}}       & 0.5 \\

\multicolumn{2}{c}{\seg{0.5}{Ba ngla deshi s}}           & 0.5    &
\multicolumn{2}{c}{\seg{0.5}{\htok{רי כוז י ים}}}  & 0.5  &
\multicolumn{2}{c}{\seg{0.4}{\ditok{डिस् ट्री ब्यू टर ों}}}  & 0.4 &
\multicolumn{2}{c}{\seg{0.0}{\tatok{தொ டுக்க}}}         & 0    &
\multicolumn{2}{c}{\seg{0.33}{zene szerz ő zs eni ket}}     & 0.33 &
\multicolumn{2}{c}{\seg{0.5}{keha bis an}}       & 0.5 \\

\multicolumn{2}{c}{\seg{0.5}{Ba ngla deshi s}}           & 0.5    &
\multicolumn{2}{c}{\seg{0.5}{\htok{רי כוז י ים}}}  & 0.5  &
\multicolumn{2}{c}{\seg{0.4}{\ditok{डिस् ट्री ब्यू टर ों}}}  & 0.4 &
\multicolumn{2}{c}{\seg{0.0}{\tatok{தொ டுக்க}}}         & 0    &
\multicolumn{2}{c}{\seg{0.33}{zene szerz ő zs eni ket}}     & 0.33 &
\multicolumn{2}{c}{\seg{0.5}{keha bis an}}       & 0.5 \\
\bottomrule

\multicolumn{3}{c}{meadowlarks} &
\multicolumn{3}{c}{\htok{גבשושיים}} &
\multicolumn{3}{c}{\ditok{रजिस्ट्रारों}} &
\multicolumn{3}{c}{\tatok{வசித்து}} &
\multicolumn{3}{c}{megtörténhetésének} &
\multicolumn{3}{c}{disucikan} \\

\multicolumn{3}{c}{meadowlark s} &
\multicolumn{3}{c}{\htok{גבשושי ים}} &
\multicolumn{3}{c}{\ditok{रजिस्ट्रार ों}} &
\multicolumn{3}{c}{\tatok{வசி த்து}} &
\multicolumn{3}{c}{megtörténhetés ének} &
\multicolumn{3}{c}{di suci kan} \\

\midrule

\multicolumn{2}{c}{\textit{Segmentation}} & \textit{F1} &
\multicolumn{2}{c}{\textit{Segmentation}} & \textit{F1} &
\multicolumn{2}{c}{\textit{Segmentation}} & \textit{F1} &
\multicolumn{2}{c}{\textit{Segmentation}} & \textit{F1} &
\multicolumn{2}{c}{\textit{Segmentation}} & \textit{F1} &
\multicolumn{2}{c}{\textit{Segmentation}} & \textit{F1} \\

\midrule

\multicolumn{2}{c}{\seg{0.0}{mea do wl arks}}          & 0   &
\multicolumn{2}{c}{\seg{0.0}{\htok{ג בשו שיים}}}  & 0   &
\multicolumn{2}{c}{\seg{0.0}{\ditok{रजि स्ट्र ारों}}}  & 0   &
\multicolumn{2}{c}{\seg{1.0}{\tatok{வசி த்து}}}       & 1   &
\multicolumn{2}{c}{\seg{0.0}{megt örtén he té sének}}    & 0   &
\multicolumn{2}{c}{\seg{0.5}{dis uci kan}}        & 0.5  \\

\multicolumn{2}{c}{\seg{0.5}{mea dow lark s}}         & 0.5 &
\multicolumn{2}{c}{\seg{0.5}{\htok{ג בש ושי ים}}} & 0.5  &
\multicolumn{2}{c}{\seg{0.5}{\ditok{रजिस् ट्रा र ों}}}  & 0.5 &
\multicolumn{2}{c}{\seg{1.0}{\tatok{வசி த்து}}}       & 1   &
\multicolumn{2}{c}{\seg{0.4}{meg törté nhet és ének}}     & 0.4  &
\multicolumn{2}{c}{\seg{0.8}{di s uci kan}}       & 0.80 \\

\multicolumn{2}{c}{\seg{0.5}{mea dow lark s}}           & 0.5    &
\multicolumn{2}{c}{\seg{0.67}{\htok{ג בש ושי ים}}} & 0.67 &
\multicolumn{2}{c}{\seg{0.0}{\ditok{रजिस् ट्रा रों}}}  & 0 &
\multicolumn{2}{c}{\seg{1.0}{\tatok{வசி த்து}}}       & 1   &
\multicolumn{2}{c}{\seg{0.0}{meg törté nhet ésé nek}}     & 0    &
\multicolumn{2}{c}{\seg{0.5}{dis uci kan}}       & 0.5 \\

\multicolumn{2}{c}{\seg{0.5}{mea dow lark s}}           & 0.5    &
\multicolumn{2}{c}{\seg{0.4}{\htok{ג בש וש י ים}}}  & 0.4  &
\multicolumn{2}{c}{\seg{0.5}{\ditok{रजिस् ट्रा र ों}}}  & 0.5 &
\multicolumn{2}{c}{\seg{0.67}{\tatok{வசி த்த ு}}}        & 0.67    &
\multicolumn{2}{c}{\seg{0.33}{meg törté n het és ének}}     & 0.33 &
\multicolumn{2}{c}{\seg{0.5}{dis uci kan}}       & 0.5 \\

\multicolumn{2}{c}{\seg{0.5}{mea dow lark s}}           & 0.5    &
\multicolumn{2}{c}{\seg{0.4}{\htok{ג בש וש י ים}}}  & 0.4  &
\multicolumn{2}{c}{\seg{0.5}{\ditok{रजिस् ट्रा र ों}}}  & 0.5 &
\multicolumn{2}{c}{\seg{0.67}{\tatok{வசி த்த ு}}}        & 0.67   &
\multicolumn{2}{c}{\seg{0.33}{meg törté n het és ének}}     & 0.33 &
\multicolumn{2}{c}{\seg{0.5}{dis uci kan}}        & 0.5  \\
\bottomrule

\multicolumn{3}{c}{eschewing} &
\multicolumn{3}{c}{\htok{מאושפזות}} &
\multicolumn{3}{c}{\ditok{पुरातत्वविदों}} &
\multicolumn{3}{c}{\tatok{எல்லைய்}} &
\multicolumn{3}{c}{szavazatából} &
\multicolumn{3}{c}{kegigihan} \\

\multicolumn{3}{c}{eschew ing} &
\multicolumn{3}{c}{\htok{מ אושפז ות}} &
\multicolumn{3}{c}{\ditok{पुरातत्वविद ों}} &
\multicolumn{3}{c}{\tatok{எல்லை ய்}} &
\multicolumn{3}{c}{szavazat ából} &
\multicolumn{3}{c}{ke gigih an} \\

\midrule

\multicolumn{2}{c}{\textit{Segmentation}} & \textit{F1} &
\multicolumn{2}{c}{\textit{Segmentation}} & \textit{F1} &
\multicolumn{2}{c}{\textit{Segmentation}} & \textit{F1} &
\multicolumn{2}{c}{\textit{Segmentation}} & \textit{F1} &
\multicolumn{2}{c}{\textit{Segmentation}} & \textit{F1} &
\multicolumn{2}{c}{\textit{Segmentation}} & \textit{F1} \\

\midrule

\multicolumn{2}{c}{\seg{0.0}{es che wing}}          & 0   &
\multicolumn{2}{c}{\seg{0.0}{\htok{מא ושפ זות}}}  & 0   &
\multicolumn{2}{c}{\seg{0.0}{\ditok{पुरा तत् ववि दों}}}  & 0   &
\multicolumn{2}{c}{\seg{1.0}{\tatok{எல்லை ய்}}}       & 1  &
\multicolumn{2}{c}{\seg{0.67}{szav azat ából}}    & 0.67   &
\multicolumn{2}{c}{\seg{0.0}{keg igi han}}        & 0   \\

\multicolumn{2}{c}{\seg{0.67}{esc hew ing}}         & 0.67 &
\multicolumn{2}{c}{\seg{0.4}{\htok{מאו שפ ז ות}}} & 0.4  &
\multicolumn{2}{c}{\seg{0.5}{\ditok{पुरा तत् वविद ों}}}  & 0.5 &
\multicolumn{2}{c}{\seg{0.67}{\tatok{எல்லை ய ்}}}         & 0.67    &
\multicolumn{2}{c}{\seg{0.67}{szava zat ából}}     & 0.67  &
\multicolumn{2}{c}{\seg{0.67}{ke gi g ih an}}       & 0.67 \\

\multicolumn{2}{c}{\seg{0.67}{esc hew ing}}           & 0.67    &
\multicolumn{2}{c}{\seg{0.67}{\htok{מ א ושפ ז ות}}} & 0.67 &
\multicolumn{2}{c}{\seg{0.4}{\ditok{पु रात त्व विद ों}}}  & 0.4 &
\multicolumn{2}{c}{\seg{0.0}{\tatok{எ ல்லைய ்}}}         & 0    &
\multicolumn{2}{c}{\seg{0.0}{sz avaz atá ból}}     & 0    &
\multicolumn{2}{c}{\seg{0.67}{ke g igi h an}}       & 0.67 \\

\multicolumn{2}{c}{\seg{0.67}{esc hew ing}}           & 0.67    &
\multicolumn{2}{c}{\seg{0.4}{\htok{מא ושפ ז ות}}}  & 0.4  &
\multicolumn{2}{c}{\seg{0.4}{\ditok{पु रात त्व विद ों}}}  & 0.4 &
\multicolumn{2}{c}{\seg{0.0}{\tatok{எ ல்லைய ்}}}         & 0    &
\multicolumn{2}{c}{\seg{0.5}{sz avaz at ából}}     & 0.5 &
\multicolumn{2}{c}{\seg{0.67}{ke g igi h an}}       & 0.67 \\

\multicolumn{2}{c}{\seg{0.67}{esc hew ing}}           & 0.67    &
\multicolumn{2}{c}{\seg{0.4}{\htok{מא ושפ ז ות}}}  & 0.4  &
\multicolumn{2}{c}{\seg{0.4}{\ditok{पु रात त्व विद ों}}}  & 0.4 &
\multicolumn{2}{c}{\seg{0.67}{\tatok{எல்லை ய ்}}}        & 0.67    &
\multicolumn{2}{c}{\seg{0.5}{sz avaz at ából}}     & 0.5 &
\multicolumn{2}{c}{\seg{0.4}{keg igi h an}}        & 0.4  \\
\bottomrule

\multicolumn{3}{c}{Veterans} &
\multicolumn{3}{c}{\htok{אידיאלים}} &
\multicolumn{3}{c}{\ditok{प्रतिष्ठानों}} &
\multicolumn{3}{c}{\tatok{படையைச்}} &
\multicolumn{3}{c}{tartozásának} &
\multicolumn{3}{c}{perumahan} \\

\multicolumn{3}{c}{Veteran s} &
\multicolumn{3}{c}{\htok{אידיאל ים}} &
\multicolumn{3}{c}{\ditok{प्रतिष्ठान ों}} &
\multicolumn{3}{c}{\tatok{படை யைச்}} &
\multicolumn{3}{c}{tartozás ának} &
\multicolumn{3}{c}{pe rumah an} \\

\midrule

\multicolumn{2}{c}{\textit{Segmentation}} & \textit{F1} &
\multicolumn{2}{c}{\textit{Segmentation}} & \textit{F1} &
\multicolumn{2}{c}{\textit{Segmentation}} & \textit{F1} &
\multicolumn{2}{c}{\textit{Segmentation}} & \textit{F1} &
\multicolumn{2}{c}{\textit{Segmentation}} & \textit{F1} &
\multicolumn{2}{c}{\textit{Segmentation}} & \textit{F1} \\

\midrule

\multicolumn{2}{c}{\seg{0.0}{Ve ter ans}}          & 0   &
\multicolumn{2}{c}{\seg{0.0}{\htok{אידי אלים}}}  & 0   &
\multicolumn{2}{c}{\seg{0.0}{\ditok{प्रति ष्ठा नों}}}  & 0   &
\multicolumn{2}{c}{\seg{1.0}{\tatok{படை யைச்}}}       & 1   &
\multicolumn{2}{c}{\seg{0.67}{tar tozás ának}}    & 0.67   &
\multicolumn{2}{c}{\seg{0.0}{per uma han}}        & 0   \\

\multicolumn{2}{c}{\seg{0.5}{V et eran s}}         & 0.5 &
\multicolumn{2}{c}{\seg{0.67}{\htok{אי דיאל ים}}} & 0.67  &
\multicolumn{2}{c}{\seg{0.5}{\ditok{प्रति ष्ठ ान ों}}}  & 0.5 &
\multicolumn{2}{c}{\seg{1.0}{\tatok{படை யைச்}}}       & 1   &
\multicolumn{2}{c}{\seg{0.67}{tarto zás ának}}     & 0.67  &
\multicolumn{2}{c}{\seg{1.0}{pe rumah an}}       & 1 \\

\multicolumn{2}{c}{\seg{0.67}{Vet eran s}}           & 0.67    &
\multicolumn{2}{c}{\seg{0.67}{\htok{אי דיאל ים}}} & 0.67 &
\multicolumn{2}{c}{\seg{0.5}{\ditok{प्रति ष्ठ ान ों}}}  & 0.5 &
\multicolumn{2}{c}{\seg{1.0}{\tatok{படை யைச்}}}       & 1   &
\multicolumn{2}{c}{\seg{0.0}{tarto zásá nak}}     & 0    &
\multicolumn{2}{c}{\seg{0.5}{per umah an}}       & 0.5 \\

\multicolumn{2}{c}{\seg{0.67}{Vet eran s}}           & 0.67    &
\multicolumn{2}{c}{\seg{0.67}{\htok{אי דיאל ים}}}  & 0.67  &
\multicolumn{2}{c}{\seg{0.5}{\ditok{प्र तिष्ठ ान ों}}}  & 0.5 &
\multicolumn{2}{c}{\seg{0.67}{\tatok{ப டை யைச்}}}        & 0.67    &
\multicolumn{2}{c}{\seg{0.67}{tarto zás ának}}     & 0.67 &
\multicolumn{2}{c}{\seg{0.5}{per umah an}}       & 0.5 \\

\multicolumn{2}{c}{\seg{0.67}{Vet eran s}}           & 0.67    &
\multicolumn{2}{c}{\seg{0.5}{\htok{אי דיאל י ם}}}  & 0.5  &
\multicolumn{2}{c}{\seg{0.5}{\ditok{प्र तिष्ठ ान ों}}}  & 0.5 &
\multicolumn{2}{c}{\seg{0.67}{\tatok{ப டை யைச்}}}        & 0.67    &
\multicolumn{2}{c}{\seg{0.67}{tarto zás ának}}     & 0.67 &
\multicolumn{2}{c}{\seg{0.5}{per umah an}}        & 0.5  \\
\bottomrule

\multicolumn{3}{c}{roasted} &
\multicolumn{3}{c}{\htok{עיתונאית}} &
\multicolumn{3}{c}{\ditok{स्लाइड्स}} &
\multicolumn{3}{c}{\tatok{முடிவெடுக்கப்}} &
\multicolumn{3}{c}{atrocitások} &
\multicolumn{3}{c}{bergabung} \\

\multicolumn{3}{c}{roast ed} &
\multicolumn{3}{c}{\htok{עיתונאי ת}} &
\multicolumn{3}{c}{\ditok{स्लाइड ्स}} &
\multicolumn{3}{c}{\tatok{முடிவெடு க்கப்}} &
\multicolumn{3}{c}{atrocitás ok} &
\multicolumn{3}{c}{ber gabung} \\

\midrule

\multicolumn{2}{c}{\textit{Segmentation}} & \textit{F1} &
\multicolumn{2}{c}{\textit{Segmentation}} & \textit{F1} &
\multicolumn{2}{c}{\textit{Segmentation}} & \textit{F1} &
\multicolumn{2}{c}{\textit{Segmentation}} & \textit{F1} &
\multicolumn{2}{c}{\textit{Segmentation}} & \textit{F1} &
\multicolumn{2}{c}{\textit{Segmentation}} & \textit{F1} \\

\midrule

\multicolumn{2}{c}{\seg{0.0}{roa sted}}          & 0   &
\multicolumn{2}{c}{\seg{0.0}{\htok{עיתונ אית}}}  & 0   &
\multicolumn{2}{c}{\seg{0.67}{\ditok{स्ला इड ्स}}}  & 0.67   &
\multicolumn{2}{c}{\seg{0.0}{\tatok{முடிவ ெடுக் கப்}}}       & 0   &
\multicolumn{2}{c}{\seg{0.0}{atr oci tások}}    & 0   &
\multicolumn{2}{c}{\seg{0.0}{berg abung}}        & 0   \\

\multicolumn{2}{c}{\seg{1.0}{roast ed}}         & 1.0 &
\multicolumn{2}{c}{\seg{0.67}{\htok{עיתונ אי ת}}} & 0.67  &
\multicolumn{2}{c}{\seg{0.67}{\ditok{स्ल ाइड ्स}}}  & 0.67 &
\multicolumn{2}{c}{\seg{0.5}{\tatok{முடிவ ெடு க்க ப்}}}         & 0.5    &
\multicolumn{2}{c}{\seg{0.4}{a tro c itás ok}}     & 0.4  &
\multicolumn{2}{c}{\seg{0.67}{ber gab ung}}       & 0.67 \\

\multicolumn{2}{c}{\seg{1.0}{roast ed}}         & 1.0 &
\multicolumn{2}{c}{\seg{0.67}{\htok{עיתונ אי ת}}} & 0.67 &
\multicolumn{2}{c}{\seg{0.5}{\ditok{स ्लाइ ड ्स}}}  & 0.5 &
\multicolumn{2}{c}{\seg{0.5}{\tatok{முடிவ ெடு க்க ப்}}}         & 0.5    &
\multicolumn{2}{c}{\seg{0.4}{a tro c itás ok}}     & 0.4    &
\multicolumn{2}{c}{\seg{0.67}{ber gab ung}}       & 0.67 \\

\multicolumn{2}{c}{\seg{1.0}{roast ed}}         & 1.0 &
\multicolumn{2}{c}{\seg{0.5}{\htok{ע יתונא י ת}}}  & 0.5  &
\multicolumn{2}{c}{\seg{0.5}{\ditok{स ्लाइ ड ्स}}}  & 0.5 &
\multicolumn{2}{c}{\seg{0.0}{\tatok{முடிவ ெ டுக்க ப்}}}        & 0    &
\multicolumn{2}{c}{\seg{0.5}{a troc itás ok}}     & 0.5 &
\multicolumn{2}{c}{\seg{0.67}{ber gab ung}}       & 0.67 \\

\multicolumn{2}{c}{\seg{1.0}{roast ed}}         & 1.0 &
\multicolumn{2}{c}{\seg{0.5}{\htok{ע יתונא י ת}}}  & 0.5  &
\multicolumn{2}{c}{\seg{0.67}{\ditok{स्ल ाइड ्स}}}  & 0.67 &
\multicolumn{2}{c}{\seg{0.0}{\tatok{முடிவ ெ டுக்க ப்}}}        & 0    &
\multicolumn{2}{c}{\seg{0.67}{atroc itás ok}}     & 0.67 &
\multicolumn{2}{c}{\seg{0.5}{b ergab ung}}        & 0.5  \\
\bottomrule

\end{tabular}
\caption{Evolution of segmentation of few more candidate wordforms produced by \textit{tokenizer-free} approach \g{SSLM} with morphological alignment F1-scores. English, Hebrew, and Hindi are fusional or analytical languages, while Tamil, Hungarian, and Indonesian are agglutinative languages.}
\label{tab:seg_f1_more}
\end{table*}

\begin{table*}[t]
\centering
\small
\begin{tabularx}{\textwidth}{l l l X l l}
\toprule
\textbf{Language} & \textbf{ISO 639-3} & \textbf{ISO 15924} & \textbf{Typology} & \textbf{Family} & \textbf{Group} \\
\midrule
\rowcolor[gray]{.95} \multicolumn{6}{l}{\textit{Agglutinative}} \\
Finnish & fin & latn & Agglutinative (with some Fusion) & Uralic & Finno-Ugric \\
Hungarian & hun & latn & Agglutinative (Suffixing) & Uralic & Finno-Ugric \\
Malayalam & mal & mlym & Agglutinative (Highly Synthetic) & Dravidian & Dravidian \\
Tamil & tam & taml & Agglutinative (Highly Synthetic) & Dravidian & Dravidian \\
Telugu & tel & telu & Agglutinative (Highly Synthetic) & Dravidian & Dravidian \\
Kyrgyz & kir & cyrl & Agglutinative (Suffixing) & Turkic & Turkic \\
Turkish & tur & latn & Agglutinative (Highly Productive Suffixing) & Turkic & Turkic \\
Mongolian & mon & cyrl & Agglutinative (Suffixing) & Mongolic & Mongolic \\
Indonesian & ind & latn & Agglutinative (Reduplication-Heavy) & Austronesian & Malayo-Polynesian \\
\midrule
\rowcolor[gray]{.95} \multicolumn{6}{l}{\textit{Fusional}} \\
Sanskrit & san & deva & Fusional (Polysynthetic Tendencies) & Indo-European & Indo-Aryan \\
Hindi & hin & deva & Fusional / Analytic (Split-Ergative) & Indo-European & Indo-Aryan \\
Sindhi & snd & arab & Fusional (Moderate) & Indo-European & Indo-Aryan \\
Croatian & hrv & latn & Fusional (Highly Synthetic) & Indo-European & Slavic \\
Russian & rus & cyrl & Fusional (Highly Synthetic) & Indo-European & Slavic \\
Persian & fas & arab & Fusional (Weakly Synthetic) & Indo-European & Indo-Iranian \\
\midrule
\rowcolor[gray]{.95} \multicolumn{6}{l}{\textit{Analytic \& Introflexive}} \\
English & eng & latn & Analytic / Weakly Fusional & Indo-European & Germanic \\
Swedish & swe & latn & Analytic / Weakly Fusional & Indo-European & Germanic \\
Hebrew & heb & hebr & Introflexive (Non-concatenative) & Afroasiatic & Semitic \\
\bottomrule
\end{tabularx}
\caption{Languages categorized by morphological typology. Grouping highlights the structural similarities in word formation across different language families.}
\label{tab:language-coverage-table}
\end{table*}

\begin{table*}[t]
\centering
\small
\begin{tabularx}{\textwidth}{l l r r r r r r l}
\toprule
\textbf{Language} & \textbf{ISO 15924} & \textbf{BP} & \textbf{\#sentences} & \textbf{\#words} & \textbf{ASL} & \textbf{\#MS} & \textbf{\#MN} & \textbf{Data Source} \\
\midrule
\rowcolor[gray]{.95} \multicolumn{9}{l}{\textit{Agglutinative}} \\
Finnish & latn & 1.0589051 & 432,422 & 4,532,824 & 10.48 & 10172 & 3,745,139 & NewsCrawl \\
Hungarian  & latn & 1.0199851 & 280,276 & 4,883,416 & 17.42 & 6350 & 1,044,996 & NewsCrawl \\
Malayalam  & mlym & 2.8852389 & 438,501 & 4,053,859 & 9.24 & 131 & - & NewsCrawl \\
Tamil  & taml & 2.7292892 & 397,206 & 4,216,313 & 10.61 & 1179 & - & NewsCrawl \\
Telugu  & telu & 2.6198705 & 475,463 & 4,582,098 & 9.64 & 8092 & - & NewsCrawl \\
Kyrgyz & cyrl & 1.963557 & 410,075 & 5,154,290 & 12.57 & 4221 & - & NewsCrawl \\
Turkish  & latn & 1.0444815 & 351,292 & 4,733,457 & 13.47 & 30076 & - & NewsCrawl \\
Mongolian  & cyrl & 1.8046135 & 623,147 & 5,719,249 & 9.18 & - & 16,221 & NLLB \\
Indonesian  & latn & 1.1788023 & 371,710 & 6,252,116 & 16.82 & 2785 & - & NewsCrawl \\
\midrule
\rowcolor[gray]{.95} \multicolumn{9}{l}{\textit{Fusional}} \\
Sanskrit  & deva & 2.5428913 & 857,298 & 5,159,696 & 6.02 & 16184 & - & NLLB \\
Hindi  & deva & 2.3701629 & 427,015 & 6,936,484 & 16.24 & 1301 & - & NewsCrawl \\
Sindhi  & arab & 1.5880165 & 614,055 & 7,258,014 & 11.82 & 3874 & - & NLLB \\
Croatian  & latn & 0.9897218 & 309,090 & 5,652,252 & 18.29 & 7749 & 1,765,011 & NewsCrawl \\
Russian  & cyrl & 1.8228284 & 341,500 & 5,296,132 & 15.51 & 21569 & 1,414,063 & NewsCrawl \\
Persian  & arab & 1.6790492 & 444,478 & 7,197,903 & 16.19 & 11859 & - & NewsCrawl \\
\midrule
\rowcolor[gray]{.95} \multicolumn{9}{l}{\textit{Analytic \& Introflexive}} \\
English  & latn & 1 & 250,000 & 6,249,150 & 25 & 3688 & 874,725 & NewsCrawl \\
Swedish  & latn & 1.0210256 & 492,345 & 6,002,549 & 12.19 & 6223 & 140,843 & NLLB \\
Hebrew  & hebr & 1.3555346 & 508,250 & 5,248,164 & 10.33 & 4641 & - & NLLB \\
\bottomrule
\end{tabularx}
\caption{Comparative Statistics after byte-premium (BP) \cite{arnett-etal-2024-a-bit-of-problem} adjustments and Dataset Sources for Languages Categorized by Morphological Typology. Abbreviations are BP: Byte-premiums, ASL: Average Sentence Length (in number of words), \#MS: Number of items in MorphScore \cite{arnett-bergen-2025-language, arnett2025evaluatingmorphologicalalignmenttokenizers}, \#MN: Number of items in MorphyNet \cite{batsuren-etal-2021-morphynet}, NLLB: No Language Left Behind \cite{nllbteam2022languageleftbehindscaling}.}
\label{tab:language-dataset-stats}
\end{table*}

\begin{table*}[t]
\centering
\small
\begin{tabular}{p{4.2cm}p{9.4cm}}
\toprule
\textbf{Method} & \textbf{Description} \\
\midrule
\rowcolor[gray]{.95} \multicolumn{2}{l}{\textit{Standard Algorithms}} \\
Byte-pair Encoding (BPE) \cite{gage_BPE_1994, sennrich-etal-2016-neural}
& Greedy frequency-based merge algorithm that iteratively combines the most frequent symbol pairs to construct a fixed subword vocabulary. \\
\\
UnigramLM \cite{kudo-2018-subword}
& Probabilistic subword model trained via EM that selects a vocabulary maximizing corpus likelihood under a mixture model. \\
\\
WordPiece \cite{wordPiece}
& Likelihood-based subword segmentation used in BERT; greedily selects merges that maximize training data likelihood. \\

\midrule
\rowcolor[gray]{.95} \multicolumn{2}{l}{\textit{Algorithmic Modifications}} \\
SaGe \cite{yehezkel-pinter-2023-incorporating}
& Subword segmentation algorithm preferring units that occur in fewer distinct contexts, encouraging morphological coherence. \\
\\
Morfessor \cite{smit-etal-2014-morfessor}
& Unsupervised morphology-inspired tokenizer that segments words into likely morpheme-like units using a probabilistic model of subword structure. \\

\midrule
\rowcolor[gray]{.95} \multicolumn{2}{l}{\textit{Inference Modifications}} \\
BPE-dropout \cite{provilkov-etal-2020-bpe}
& Subword regularization method that introduces stochasticity into the deterministic Byte Pair Encoding process by randomly dropping merge operations during training, allowing a model to observe multiple possible segmentations of the same word while remaining fully compatible with standard \ora{BPE} at inference time. \\
\\
PathPiece \cite{schmidt-etal-2024-tokenization-is-more-than-compression}
& Lossless subword tokenizer that utilizes a directed acyclic graph (DAG) to find the shortest path of tokens from a given vocabulary, thereby segmenting a document into the minimum possible number of tokens. \\

\midrule
\rowcolor[gray]{.95} \multicolumn{2}{l}{\textit{Vocabulary Modifications}} \\
PickyBPE \cite{chizhov-etal-2024-bpe}
& Performs vocabulary refinement during training by using the Intersection over Self (IoS) metric to identify and remove intermediate ``junk'' tokens, thereby improving vocabulary efficiency and reducing under-trained tokens without compromising text compression. \\

\midrule
\rowcolor[gray]{.95} \multicolumn{2}{l}{\textit{Pre-tokenization Modifications}} \\

MorphBPE \cite{banerjee-bhattacharyya-2018-meaningless}, MorphULM, MorphWP & Incorporates morphological boundary detection into the pre-tokenization stage prior to BPE learning, and similarly to UnigramLM (i.e., MorphULM) and WordPiece learning (i.e., MorphWP). \\

\midrule
\rowcolor[gray]{.95} \multicolumn{2}{l}{\textit{Encoding Modifications}} \\
MYTE \cite{limisiewicz-etal-2024-myte} & Alternative byte-level encoding replacing standard UTF-8 handling to better structure multilingual character representation. \\

\midrule
\rowcolor[gray]{.95} \multicolumn{2}{l}{\textit{Cross-token Algorithms}} \\
BoundlessBPE \cite{schmidt2025boundlessbytepairencoding}
& Removes explicit word-boundary constraints, allowing merges across whitespace-defined token limits. \\
\\
SuperBPE \cite{liu2025superbpespacetravellanguage}
& Extends BPE to permit cross-boundary merges and improved vocabulary utilization across token splits. \\


\bottomrule
\end{tabular}
\caption{List of Fixed-tokenizer approaches included in this study, organized by the stage at which they modify the tokenization pipeline.}
\label{tab:tokenizer_taxonomy}
\end{table*}


\begin{table*}[t]
    \centering
    \small
    \begin{tabularx}{\textwidth}{>{\raggedright\arraybackslash}X >{\raggedright\arraybackslash}X >{\raggedright\arraybackslash}X >{\raggedright\arraybackslash}X}
        \toprule
        \textbf{Metric} & \textbf{Description} & \textbf{Requirements} & \textbf{What it Addresses} \\
        \midrule
        
        \rowcolor[gray]{.95} \multicolumn{4}{l}{\textit{Coverage}} \\
        Subword Fertility & Average number of tokens produced per word. & Tokenizer and representative corpus. & Sequence length and basic storage efficiency. \\
        Compression Rate & Ratio of raw text size to tokenized sequence length. & Tokenizer and representative corpus. & Storage and context window efficiency. \\
        Token Frequency Distribution & Distributional profile of token occurrences (e.g., long-tail vs. concentrated usage). & Token counts from a representative corpus. & Statistical balance of vocabulary usage and fragmentation patterns. \\
        Token Length Distribution & Distributional profile of tokens grouped by lengths from a representative corpus & Vocabulary of tokenizer & Granularity balance between short and long tokens, which affects efficiency and segmentation behavior. \\
        Contextual Exponence & Number of distinct neighbours each token encounters \cite{yehezkel-pinter-2023-incorporating}. We consider only immediate neighbourhoods in our evaluation. & Representative corpus and its corresponding tokens & Degree to which the tokenizer optimizes tokens' contextual soundness \\
        Effective Vocabulary Size & Total number of unique tokens available in the tokenizer vocabulary. For \textit{tokenizer-free} approaches, we consider number of unique tokens when a corpus is segmented using it. & Trained tokenizer specification or model. & Memory footprint, embedding-table size, and granularity trade-offs. \\
        \midrule
        \rowcolor[gray]{.95} \multicolumn{4}{l}{\textit{Generalizability}} \\
        Type-Token Ratio (TTR) & Ratio of unique token types to total token occurrences in a corpus sample. & Tokenized corpus with fixed sample size (or moving-window normalization). & Lexical diversity and degree of repetition; helps compare whether a tokenizer yields broader vs. concentrated token usage. \\
        \midrule
        \rowcolor[gray]{.95} \multicolumn{4}{l}{\textit{Linguistic Alignment}} \\
        Morphological Alignment & Divergence between token segments and known morphemes. & Morphological analyzer and gold-standard data. & Linguistic fidelity and grammatical preservation. \\
        \midrule
        \rowcolor[gray]{.95} \multicolumn{4}{l}{\textit{Robustness}} \\
        Rényi Efficiency \cite{zouhar-etal-2023-tokenization-and-noiseless-channel} & Information-theoretic metric penalizing skewed distributions. & Token frequency data from a corpus. & Statistical balance and information density. \\
        \bottomrule
    \end{tabularx}
    \caption{Summary of Tokenizer Evaluation Metrics grouped by the proposed multidimensional framework in \citet{alqahtani2026stoptakingtokenizersgranted}}
    \label{tab:tokenizer-metrics}
\end{table*}

\begin{figure}[t]
  \centering
  \includegraphics[width=\columnwidth]{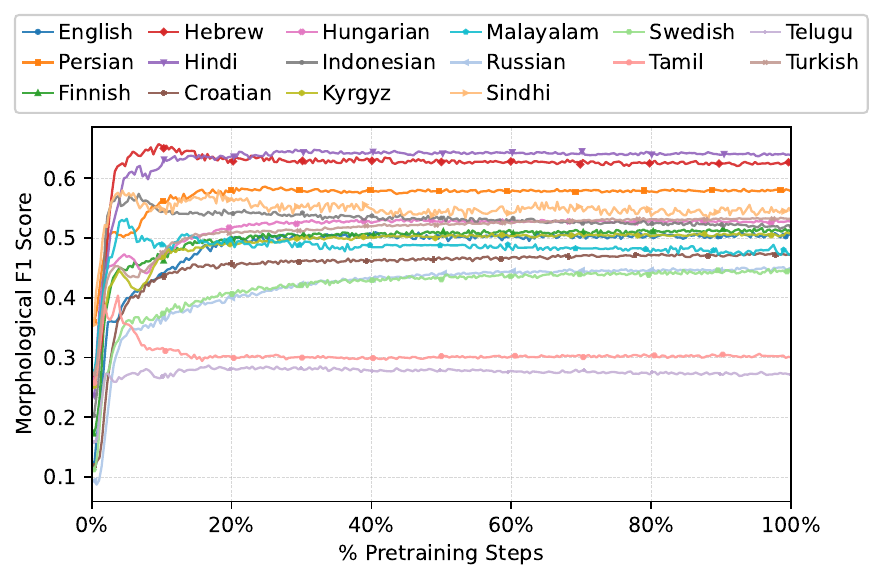}
  \caption{Evolution of Morphological Alignment (F1-score) in Tranformer-based Subword Segmental Language Models (\g{SSLM}).}
  \label{fig:morphscore_dynamics_all_languages}
\end{figure}

\begin{figure*}[t]
  \centering
  \begin{subfigure}[t]{0.48\textwidth}
    \centering
    \includegraphics[width=\linewidth]{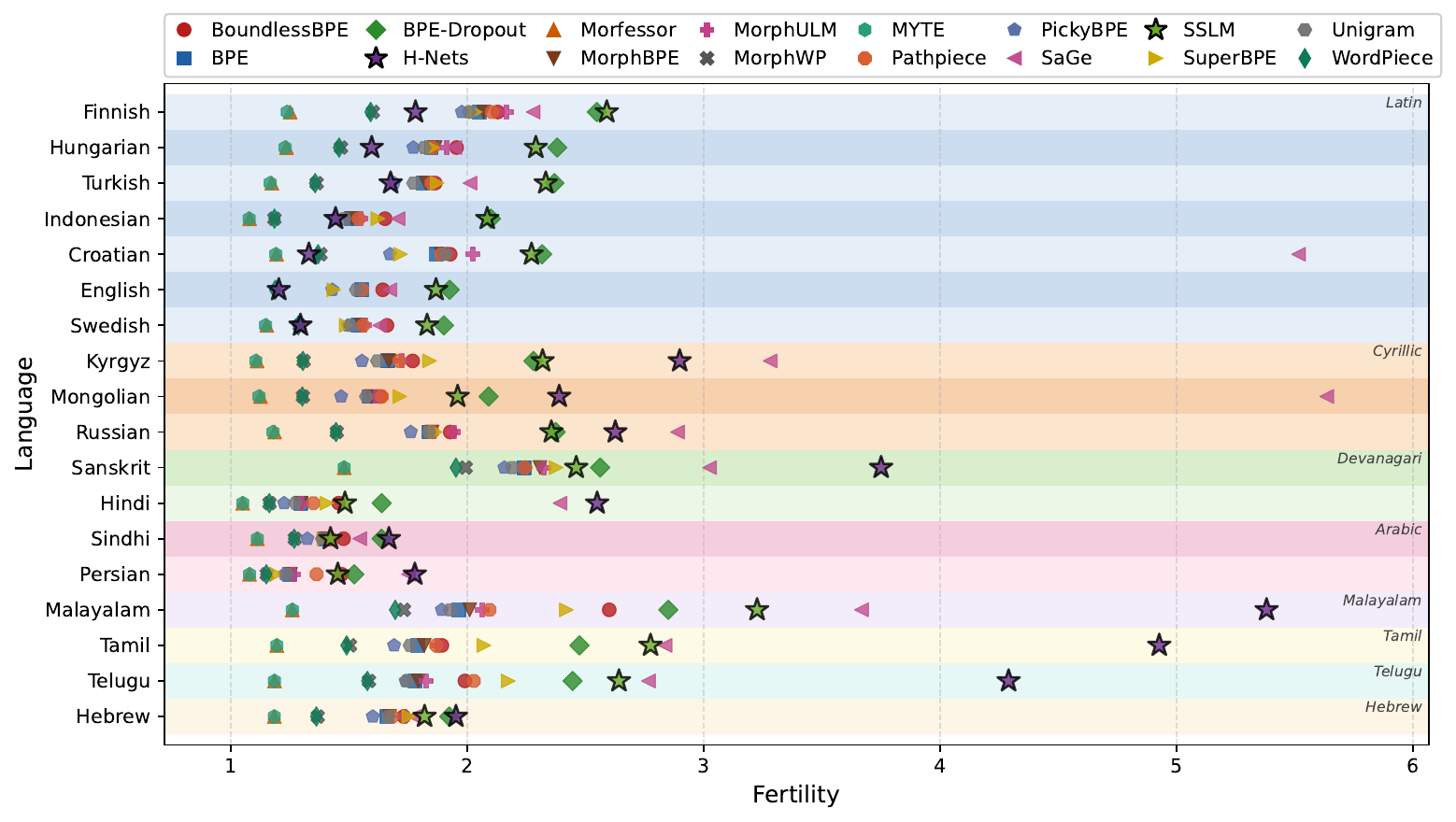}
    \caption{Fertility by script}
    \label{fig:script_fertility}
  \end{subfigure}
  \hfill
  \begin{subfigure}[t]{0.48\textwidth}
    \centering
    \includegraphics[width=\linewidth]{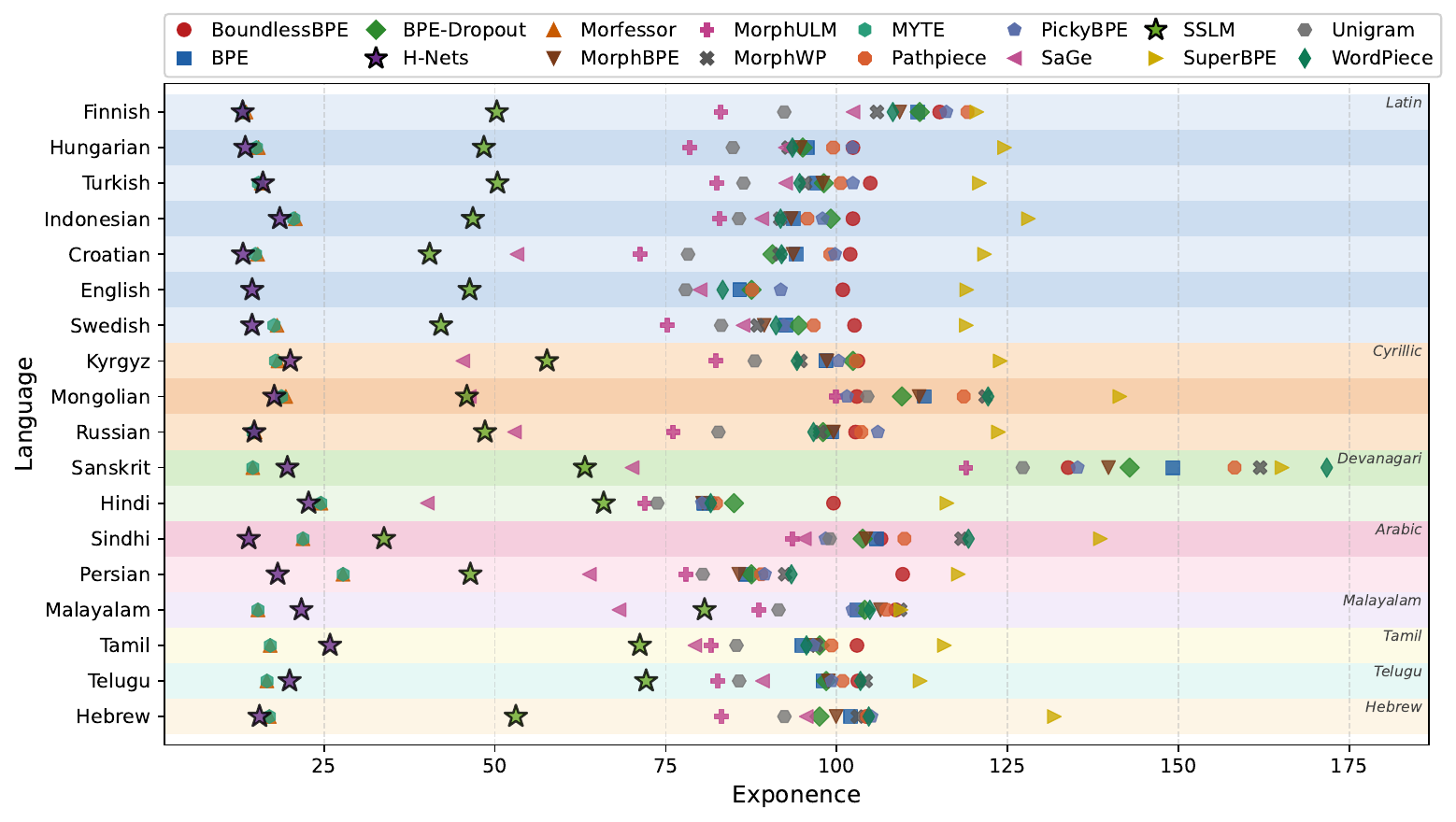}
    \caption{Contextual exponence by script}
    \label{fig:script_exponence}
  \end{subfigure}

  \vspace{0.5em}

  \begin{subfigure}[t]{0.48\textwidth}
    \centering
    \includegraphics[width=\linewidth]{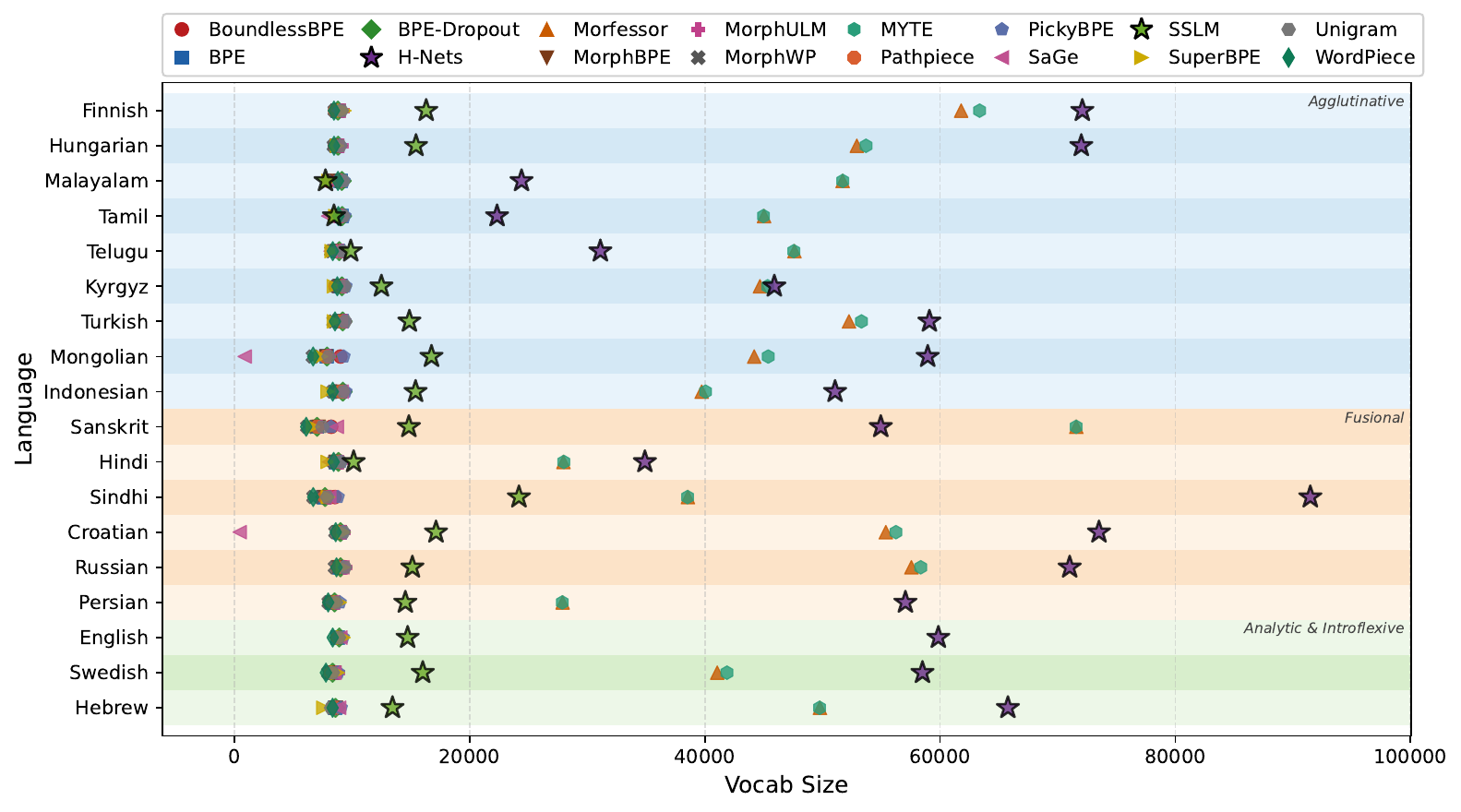}
    \caption{Effective vocabulary size by typology}
    \label{fig:typology_vocab_size}
  \end{subfigure}
  \hfill
  \begin{subfigure}[t]{0.48\textwidth}
    \centering
    \includegraphics[width=\linewidth]{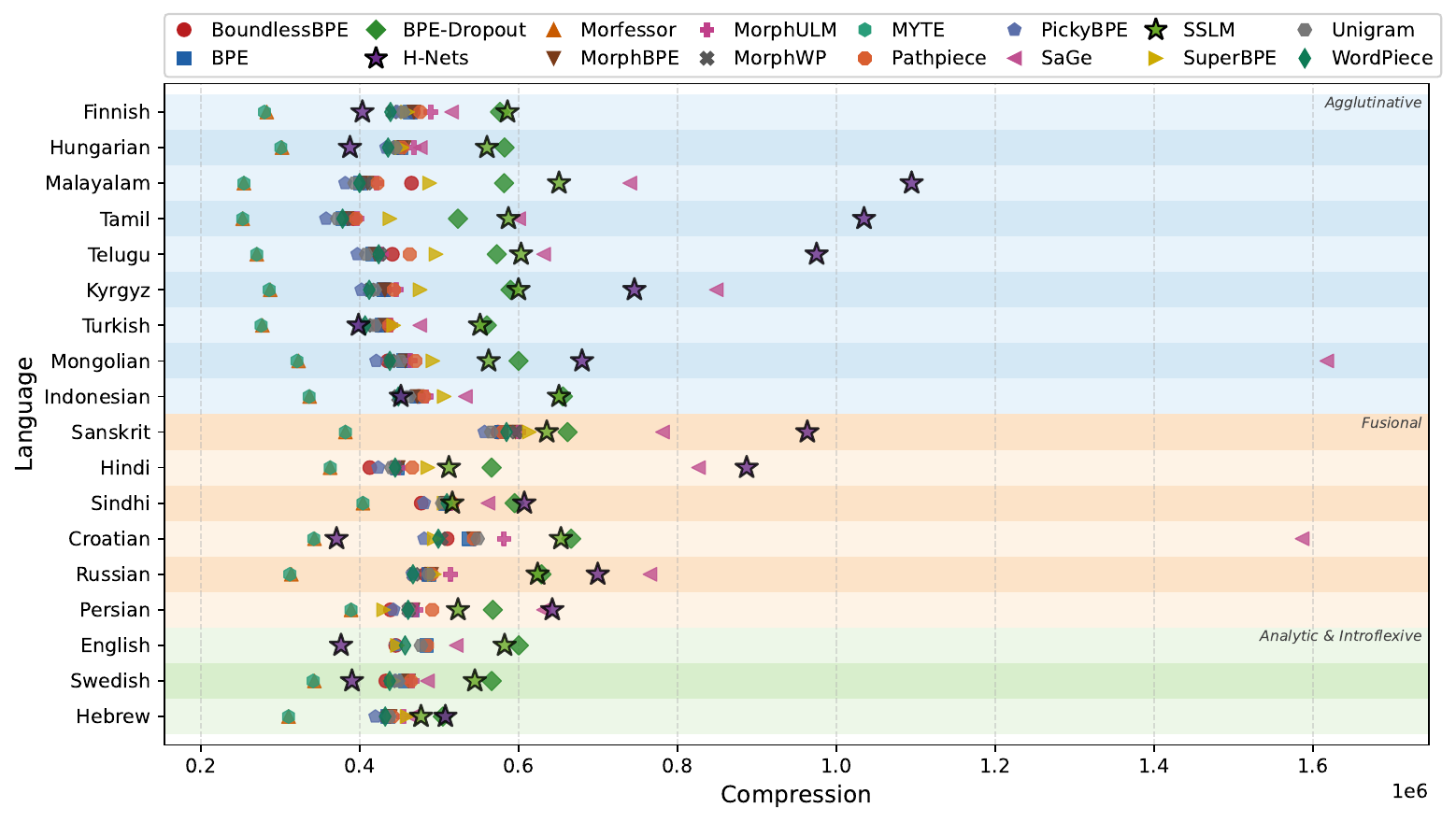}
    \caption{Compression by typology}
    \label{fig:typology_compression}
  \end{subfigure}

  \vspace{0.5em}

  \begin{subfigure}[t]{0.48\textwidth}
    \centering
    \includegraphics[width=\linewidth]{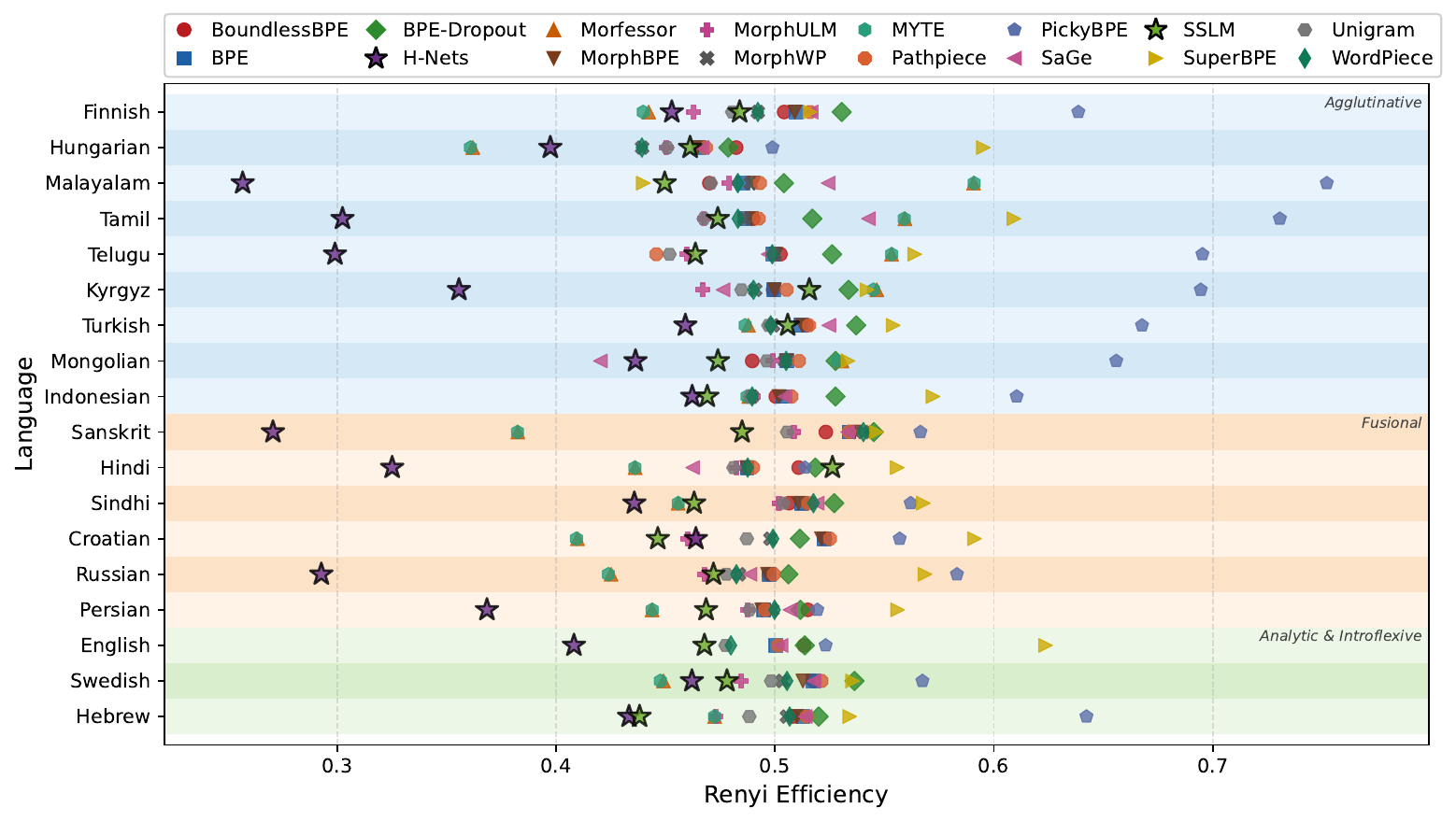}
    \caption{Rényi efficiency by typology}
    \label{fig:typology_renyi_entropy}
  \end{subfigure}
  \hfill
  \begin{subfigure}[t]{0.48\textwidth}
    \centering
    \includegraphics[width=\linewidth]{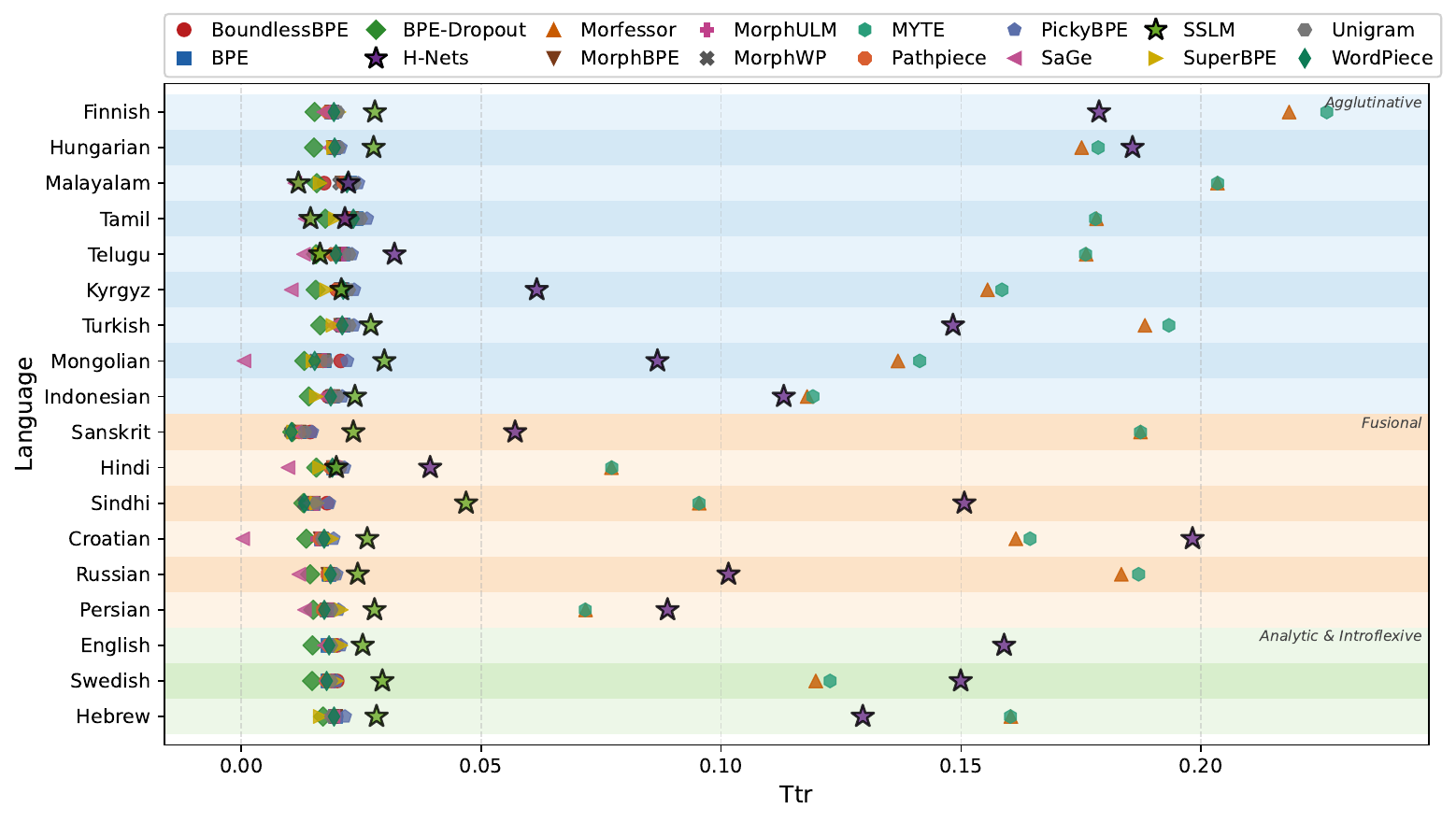}
    \caption{Type-token ratio by typology}
    \label{fig:typology_ttr}
  \end{subfigure}

  \caption{Intrinsic properties of \textit{tokenizer-free} approaches (\bl{H-Net}s and \g{SSLM}s) compared to fixed-tokenizer approaches across scripts and language typologies.}
  \label{fig:intrinsic_properties_combined}
\end{figure*}

\end{document}